%% file: article.tex
\documentclass[onefignum,onetabnum]{siamart171218}

\input{0_Title}
\usepackage{bm}
\usepackage[export]{adjustbox}
\usepackage{subfigure}
\usepackage{bold-extra}
\usepackage{booktabs}
\usepackage{mathtools}
\usepackage{caption}

\graphicspath{{./figs/}}

\ifpdf
\hypersetup{
  pdftitle={DUE},
  pdfauthor={J. Chen, K. Wu, D. Xiu}
}
\fi

\begin{document}

\maketitle

\begin{abstract}
	Equations, particularly differential equations, are fundamental for understanding natural phenomena and predicting complex dynamics across various scientific and engineering disciplines. However, the governing equations for many complex systems remain {\em unknown} due to intricate underlying mechanisms. Recent advancements in machine learning and data science offer a new paradigm for modeling unknown equations from measurement or simulation data. This paradigm shift, known as data-driven discovery or modeling, stands at the forefront of artificial intelligence for science (AI4Science), with significant progress made in recent years. In this paper, we introduce a systematic framework for data-driven modeling of unknown equations using deep learning. This versatile framework is capable of learning unknown ordinary differential equations (ODEs), partial differential equations (PDEs), differential-algebraic equations (DAEs), integro-differential equations (IDEs), stochastic differential equations (SDEs), reduced or partially observed systems, and non-autonomous differential equations. Based on this framework, we have developed Deep Unknown Equations (DUE), an open-source software package designed to facilitate the data-driven modeling of unknown equations using modern deep learning techniques. DUE serves as an educational tool for classroom instruction, enabling students and newcomers to gain hands-on experience with differential equations, data-driven modeling, and contemporary deep learning approaches such as fully connected neural networks (FNN), residual neural networks (ResNet), generalized ResNet (gResNet), operator semigroup networks (OSG-Net), and Transformers from large language models (LLMs). Additionally, DUE is a versatile and accessible toolkit for researchers across various scientific and engineering fields. It is applicable not only for learning unknown equations from data but also for surrogate modeling of known, yet complex, equations that are costly to solve using traditional numerical methods. We provide detailed descriptions of DUE and demonstrate its capabilities through diverse examples, which serve as templates that can be easily adapted for other applications. The source code for DUE is available at \url{https://github.com/AI4Equations/due}.

\end{abstract}

\begin{keywords}
education software, differential equations, deep learning,
 neural networks
\end{keywords}

\begin{AMS}
68T07, 65-01, 65-04, 37M99, 65M99, 65P99
\end{AMS}

\input{1_Introduction}

\input{2_FML}

\input{3_Usage}
\input{4_Examples}

\input{5_Conclusion}

\vspace{-2mm}

\bibliographystyle{siamplain}
\bibliography{due}

\end{document}

%% file: 0_Title.tex
\usepackage{lipsum}
\usepackage{amsfonts}
\usepackage{graphicx}
\usepackage{epstopdf}
\usepackage{algorithm}
\usepackage{algpseudocode}
\ifpdf
  \DeclareGraphicsExtensions{.eps,.pdf,.png,.jpg}
\else
  \DeclareGraphicsExtensions{.eps}
\fi

\newsiamremark{remark}{Remark}
\newsiamremark{hypothesis}{Hypothesis}
\crefname{hypothesis}{Hypothesis}{Hypotheses}
\newsiamthm{claim}{Claim}

\headers{DUE}{J. Chen, K. Wu, D. Xiu}

\title{DUE: A Deep Learning Framework and Library for Modeling\\ Unknown Equations	\thanks{
		J.~Chen and K.~Wu were partially supported by NSFC grants (No.~92370108 and No.~12171227) and Shenzhen Science and Technology Program (No.~RCJC20221008092757098).}}

\author{Junfeng Chen\thanks{Department of Mathematics and Shenzhen International Center for Mathematics, Southern University of Science and Technology, Shenzhen 518055, China (\email{chenjf2@sustech.edu.cn}).}
	\and Kailiang Wu\thanks{Corresponding author. Department of Mathematics and Shenzhen International Center for Mathematics, Southern University of Science and Technology, Shenzhen, Guangdong 518055, China (\email{wukl@sustech.edu.cn}).}
	\and Dongbin Xiu\thanks{Department of Mathematics, The Ohio State University, Columbus, OH 43210, USA (\email{xiu.16@osu.edu}).}}

\usepackage{amsopn}

\DeclareMathOperator*{\argmin}{arg\,min}

\DeclareMathAlphabet{\pazocal}{OMS}{zplm}{m}{n}

\usepackage{multirow} 

\usepackage{listings}
\usepackage{color}

\definecolor{dkgreen}{rgb}{0,0.6,0}
\definecolor{gray}{rgb}{0.5,0.5,0.5}
\definecolor{mauve}{rgb}{0.58,0,0.82}

%% file: 1_Introduction.tex
\section{Introduction}
\label{sec:intro}

Equations, especially differential equations, form the foundation of our understanding of many fundamental laws.  They help human unlock the mysteries of microscopic particles, decipher the motion of celestial bodies, predict climate changes, and explore the origins of the universe. 
Differential equations have widespread applications across disciplines such as physics, chemistry, biology, and epidemiology. Traditionally, these equations were derived from first principles. However, for many complex systems, the governing equations remain elusive due to intricate underlying mechanisms.

Recent advancements in machine learning and data science are revolutionizing how we model dynamics governed by unknown equations. This paradigm shift, known as {\em data-driven discovery or modeling}, stands at the forefront of artificial intelligence for science (AI4Science). In the past few years, significant progress has been made in learning or discovering unknown equations from data. 
Techniques such as symbolic regression \cite{bongard2007automated,schmidt2009distilling}, sparsity-promoting regression \cite{candes2006stable,wang2011predicting,tran2017exact,brunton2016discovering,rudy2017data,schaeffer2017learning,mangan2017model,brunton2017chaos}, Gaussian processes \cite{raissi2017machine},  polynomial approximation \cite{wu2019numerical,wu2020structure,ahmadi2023learning}, linear multistep methods \cite{keller2021discovery,du2022discovery}, genetic algorithms \cite{xu2020dlga,xu2021robust,chen2022symbolic}, parameter identification \cite{miao2011identifiability}, deep neural networks (DNNs) \cite{raissi2018deep,raissi2018multistep,long2018pde,long2019pde,sun2020neupde}, and neural ordinary differential equations (ODEs) \cite{chen2018neural,kim2021stiff} have shown great promise. Successfully learning these equations enables their solution using appropriate numerical schemes to predict the evolution behavior of complex systems.

 A distinct approach is using data-driven methods to learn the dynamics or flow maps of the underlying unknown equations \cite{qin2019data,wu2020data,chen2022deep}. This approach facilitates recursive predictions of a system's evolution, thereby circumventing the need to solve the learned equations. A classic example is dynamic mode decomposition (DMD) \cite{schmid2010dynamic,tu2014dynamic}, which seeks the best-fit linear operator to advance state variables forward in time, serving as an approximation to the Koopman operator associated with the underlying system \cite{brunton2022modern}. 
With the rapid development of deep learning \cite{higham2019deep}, DNNs have shown great promise in data-driven modeling of unknown equations. Compared to traditional methods, DNNs excel in managing high-dimensional problems, processing very large datasets, and facilitating parallel computing. DNNs have proven highly effective in learning the dynamics or flow maps of various types of equations, including ODEs \cite{qin2019data}, partial differential equations (PDEs) \cite{wu2020data},  differential-algebraic equations (DAEs) \cite{chen2021generalized}, integro-differential equations (IDEs) \cite{chen2022deep}, and stochastic differential equations (SDEs) \cite{chen2024learning}. This flow map learning (FML) methodology has also been extended to partially observed systems with missing state variables \cite{fu2020learning} and non-autonomous dynamical systems \cite{qin2021data}. Recent progresses in scientific machine learning (SciML) have introduced advanced deep learning techniques for approximating general operators mapping between two infinite-dimensional function spaces. Notable contributions include neural operators \cite{li2020multipole,li2021fourier,kovachki2023neural} and deep operator networks (DeepONet) \cite{lu2021learning,zou2024neuraluq}, which can also model PDEs.

\vspace{-3.999mm}

\begin{figure}[!tbhp]    
\centering
    \includegraphics[width=.93\linewidth]{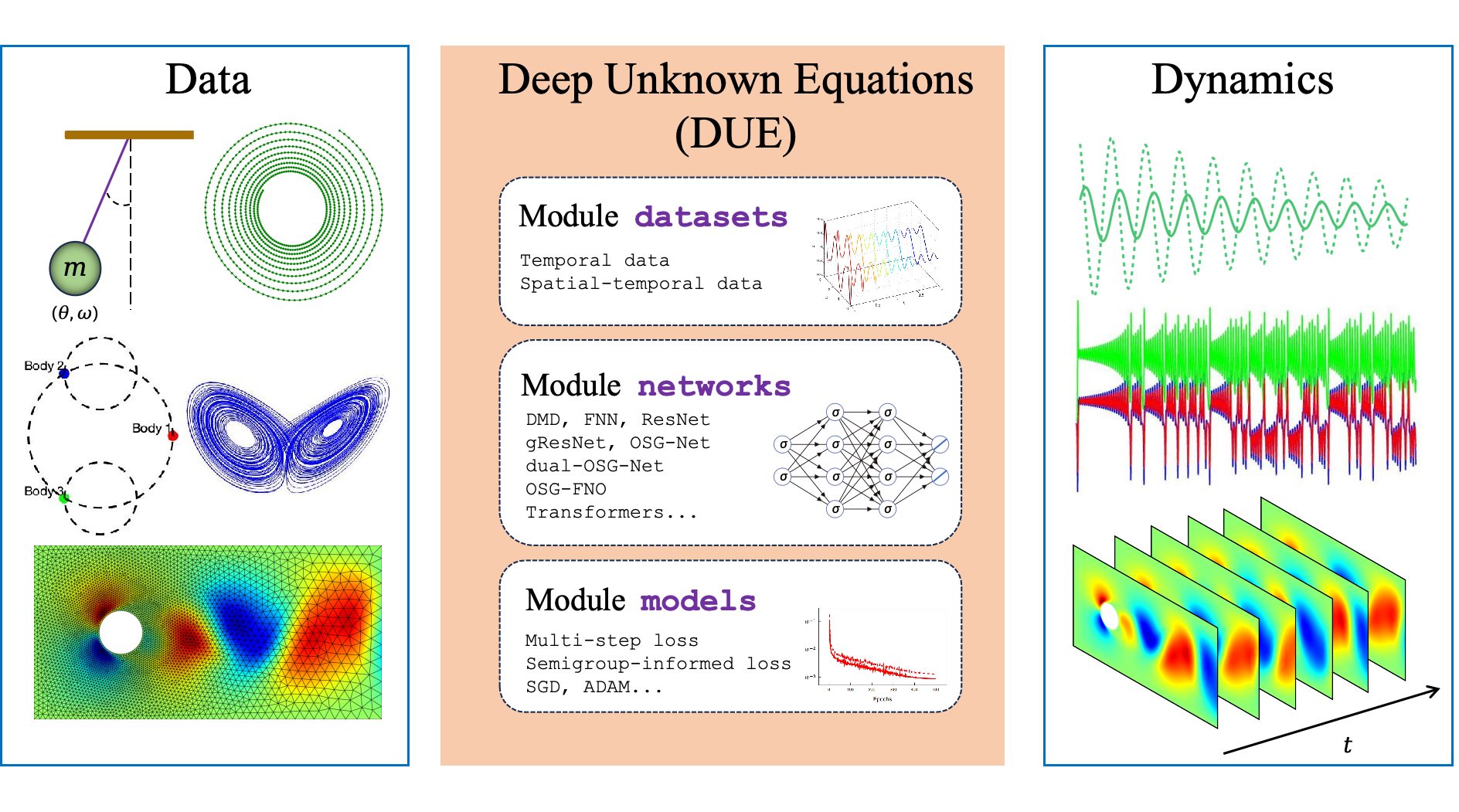}%
    \caption{The overall structure of DUE.}
    \label{fig:due}
\end{figure}

While deep learning garners growing interest among students and researchers across various fields, newcomers often encounter challenges due to the complexity of new concepts, algorithms, and coding requirements. To address this, we present Deep Unknown Equations (DUE), a  framework and open-source Python library for deep learning of unknown equations. DUE aims to simplify the learning process and facilitate the adoption of advanced deep learning techniques, such as residual neural networks (ResNet) \cite{he2016deep}, generalized ResNet (gResNet) \cite{chen2021generalized}, operator semigroup network (OSG-Net) \cite{chen2023deep}, and Transformers \cite{vaswani2017attention}. It serves as both an educational tool for students and a powerful resource for researchers, enabling the learning and modeling of any time-dependent differential equations. 
One of DUE's standout features is its user-friendly design, which allows users to start learning unknown equations with as few as {\bf ten lines of code}. This simplicity saves significant time on conceptualization and analysis, making advanced techniques more accessible. Moreover, DUE is not only valuable for learning unknown equations but also for creating surrogate models of known yet complex equations that are computationally expensive to solve using traditional numerical methods.  
As the field of deep learning continues to advance rapidly, we are committed to maintaining and updating DUE to ensure it remains a valuable tool for those interested in the deep learning of unknown equations. While similar efforts, such as DeepXDE \cite{lu2021deepxde} and NeuralUQ \cite{zou2024neuraluq}, have been made to ease the learning and adoption curve, they focus primarily on solving given differential equations or uncertainty quantification. In contrast, DUE uniquely targets the deep learning of unknown equations. 
In summary, DUE is a comprehensive framework and accessible tool that empowers students and researchers to harness deep learning for modeling unknown equations, opening new avenues in scientific discovery.

%% file: 2_FML.tex
\section{Data-Driven Deep Learning of Unknown Equations}
\label{sec:fml}
In this section, we explore how deep learning can be applied to model unknown differential equations from measurement data. After establishing the basic setup in Section \ref{sec:setup}, we introduce the essential concepts and methods for modeling unknown ODEs. This includes discussions on data preprocessing, neural network architectures, and model training, which form the core components of the deep-learning-based modeling framework. We then describe how this approach can be extended to partially observed systems. Finally, we discuss learning unknown PDEs in both nodal and modal spaces.

\subsection{Setup and Preliminaries}
\label{sec:setup}

To set the stage for our exploration, let us delve into the setup for modeling unknown ODEs \cite{qin2019data} and PDEs \cite{wu2020data,chen2022deep}. The framework we describe can be easily adapted to other types of equations, including 
 DAEs \cite{chen2021generalized}, IDEs \cite{chen2022deep}, and SDEs \cite{chen2024learning}. 

\textbf{Learning ODEs.} Imagine we are trying to understand an autonomous system where the underlying equations are {\em unknown} ODEs:
\begin{equation}\label{eq:ode}
    \frac{d\textbf{u}}{dt} = \textbf{f}(\textbf{u}(t)),\quad \textbf{u}(t_0)=\textbf{u}_0,
\end{equation}
where $\textbf{f}:\mathbb{R}^n\rightarrow\mathbb{R}^n$ is unknown. A classic example is the damped pendulum system:
\begin{equation}\label{dampedPS}
	\begin{dcases}
		\frac{d u_1}{dt}=u_2,\\
		\frac{d u_2}{dt}=-\alpha u_2- \beta \text{sin}(u_1),
	\end{dcases}
\end{equation}
where $u_1$ is the angle, $u_2$ is the angular velocity, $\alpha$ is the damping coefficient, and $\beta$ represents the effect of gravity. If these equations are known, then numerical methods like the Runge--Kutta can solve them, predicting how $u_1$ and $u_2$ evolve over time. But what if these equations are unknown? If we can observe or measure the state variables, can we build a data-driven model to predict their evolution?

Assume we have measurement data of $\textbf{u}$ collected from various trajectories. Let $t_0<t_1^{(i)}< \dots < t_K^{(i)}$ be a sequence of time instances. We use 
\begin{equation}\label{eq:ode_data}
	\textbf{u}_k^{(i)} = \textbf{u}( t_k^{(i)}; \textbf{u}_{0}^{(i)}, t_0) + {\bm \epsilon}_{\textbf{u},k}^{(i)}, \qquad k=1,\dots,K_i, \quad i=1,\dots,{I_{traj}},
\end{equation}
to denote the state at time $t_k^{(i)}$ along the $i$-th trajectory originating from the initial state $\textbf{u}_{0}^{(i)}$ at $t_0$, for a total of $I_{traj}$ trajectories. In real-world scenarios, the data may contain measurement noise ${\bm \epsilon}_{\textbf{u},k}^{(i)}$, typically modeled as random variables. Our objective is to create a data-driven model for the unknown ODEs that can predict the evolution of $\textbf{u}$ from any initial state $\textbf{u}(t_0)=\textbf{u}_0$.

\begin{figure}[tbhp]
	\centering
	\includegraphics[width=0.46\textwidth, valign=c]{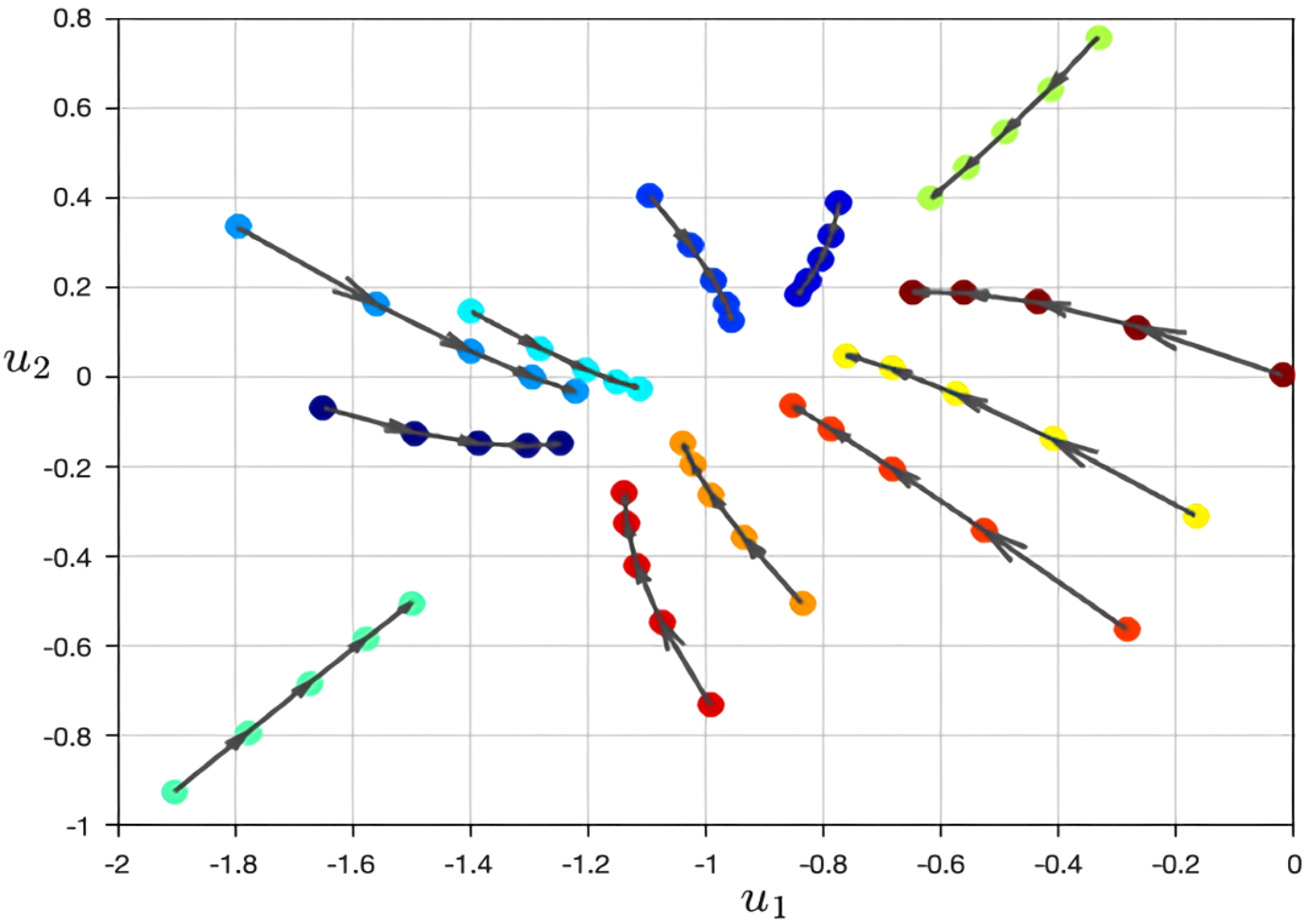} \qquad 
	\includegraphics[width=0.39\textwidth, valign=c]{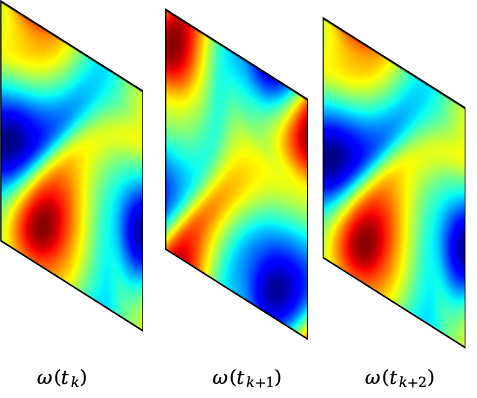}
	    \caption{Left: Trajectory data collected from multiple initial states for learning ODEs. Right: Snapshot data for learning PDEs (only one trajectory is displayed for visualization, while the real dataset may contain multiple trajectories).}
	\label{fig:data}
\end{figure}

\textbf{Learning PDEs.} Now, consider the more complex scenario of an unknown time-dependent PDE system:
\begin{equation}
	\begin{cases}
		\partial_t {\bf u}  = \mathcal{L}({\bf u}), & (x,t)\in \Omega\times \mathbb{R}^+,\\
		\mathcal{B}({\bf u}) = 0, & (x,t)\in \partial\Omega\times \mathbb{R}^+,\\
		{\bf u}(x,0)   = {\bf u}_0(x), & x\in \bar{\Omega},
	\end{cases}
	\label{eq:eg_vbg}
\end{equation}
where $\Omega\subseteq \mathbb{R}^d$ is the physical domain, $\mathcal{L}$ is the unknown operator governing the PDE, $\mathcal{B}$ specifies the boundary conditions, and the solution ${\bf u}(x,t)$ belongs to an infinite-dimensional Hilbert space $\mathbb{V}$.
A fundamental example of PDEs is the {one-dimensional} Burgers' equation:
\begin{equation*}
	\partial_t u = \mathcal{L}(u) \quad \text{with} \quad \mathcal{L}(u)=-\partial_x \left(\frac{u^2}{2}\right) + \nu \partial_{xx}u,
\end{equation*}
where the state of $u(x,t)$ is governed by a convective term $\partial_x (u^2/2)$ and a diffusive term $\nu \partial_{xx}u$, with $\nu>0$ being the diffusion coefficient (or kinematic viscosity in the context of fluid mechanics). With given initial conditions, numerical methods can predict future solutions. But what if the underlying mechanism is unclear and the right-hand side of the PDE is unknown? Can we use measurable data of $u(x,t)$ to uncover the dynamics?

Assume the solution ${\bf u}(x,t)$ of the unknown system \eqref{eq:eg_vbg} is measurable, i.e., the snapshot data of $\bf u$ are available at certain time instances as shown in \Cref{fig:data}:
\begin{equation}\label{eq:pde_data_nodal_continu}
    {\bf u}(x_s,t_k^{(i)})\qquad s=1,2\dots,n,  \quad k=1,\dots,K_i,  \quad i=1,\dots, I_{traj}.
\end{equation}
Here, $\{x_s\}_{s=1}^n$ are the discrete spatial locations at which the solution data is measured. In practice, solution data may be collected on varying sets of sampling locations, necessitating interpolation or fitting methods to transform the data onto a consistent set $\{x_s\}_{s=1}^n$. Our goal is to create a data-driven model for the unknown PDE that can predict the temporal evolution of $\bf u$ given any initial state ${\bf u}(x, 0)={\bf u}_0(x)$.

\subsection{Data Pairs}
\label{sec:set up}

In DUE, we mainly focus on learning the integral form of the underlying equations, which is equivalent to learning
the flow maps $\{\Phi_{\Delta}\}_{\Delta\geq 0}$ that describe the time evolution of state variables. 
The flow map for a time step $\Delta$ is defined as 
\begin{equation}\label{eq:ode_fmo}
	\Phi_{\Delta}(\textbf{u}_0):=\textbf{u}(t_0+\Delta) = \textbf{u}_0+\int_{t_0}^{t_0+\Delta} \textbf{f}(\textbf{u}(s)){\rm d}s = \textbf{u}_0+\int_0^{\Delta} \textbf{f}(\Phi_s(\textbf{u}_0)){\rm d}s,
\end{equation}
where $t_0$ can be arbitrarily shifted for autonomous systems. The flow maps fully characterize the system's time evolution. 
The data \eqref{eq:ode_data} may be collected at constant or varying time lags 
$\Delta_k^{(i)}=t^{(i)}_{k+1}-t^{(i)}_k$. 
Depending on this, we rearrange the data as follows:

\textbf{Rearranging Data with Fixed Time Lag \(\Delta\).} 
When data is collected at a constant time lag \(\Delta\), our goal is to learn a single flow map for this specific \(\Delta\). We segment the collected trajectories to form a dataset of input-output pairs:
\begin{equation}\label{eq:data_fixed}
	\left\{\textbf{u}_{\text{in}}^{(j)}, \textbf{u}_{\text{out}}^{(j)}\right\},\quad j= 1,2,...,J,
\end{equation}
where \(\textbf{u}_{\text{in}}^{(j)}\) and \(\textbf{u}_{\text{out}}^{(j)}\) are neighboring states such that \(\textbf{u}_{\text{out}}^{(j)} \approx \Phi_{\Delta}(\textbf{u}_{\text{in}}^{(j)})\), accounting for some measurement noise. Note that multiple data pairs can be extracted from a single trajectory by segmenting it into smaller temporal intervals, leading to $J\geq I_{traj}$.

\textbf{Rearranging Data with Varying Time Lags.} 
When the time lag \(\Delta\) varies, each \(\Delta\) represents a different flow map. Our objective becomes learning a family of flow maps \(\{\Phi_{\Delta}\}_{\Delta_1 \leq \Delta \leq \Delta_2}\), where \(\Delta_1\) and \(\Delta_2\) are the minimum and maximum time lags in the dataset. We rearrange the data into:
\begin{equation}\label{eq:data_varied}
	\left\{\textbf{u}_{\text{in}}^{(j)}, \Delta^{(j)}, \textbf{u}_{\text{out}}^{(j)}\right\},\quad j= 1,2,...,J,
\end{equation}
with \(\textbf{u}_{\text{out}}^{(j)} \approx \Phi_{\Delta^{(j)}}(\textbf{u}_{\text{in}}^{(j)})\), considering some measurement noise.

\subsection{Deep Neural Networks}
\label{sec:dnn}

In this subsection, we introduce several effective DNN architectures for modeling unknown equations, including 
the basic feedforward neural networks (FNNs), residual neural network (ResNet) \cite{he2016deep}, generalized ResNet (gResNet) \cite{chen2021generalized}, and operator semigroup network (OSG-Net) \cite{chen2023deep}.

\textbf{FNN.} As a foundational architecture in deep learning, FNN with $L$ hidden layers can be mathematically represented as:
\begin{equation}\label{eq:fnn}
	{\mathcal N}_{\bm{\theta}}(\textbf{u}_{\text{in}}) = {\bf W}_{L+1} \circ ( \sigma_L \circ  {\bf W}_{L} ) \circ \cdots \circ ( \sigma_1 \circ  {\bf W}_1 ) (\textbf{u}_{\text{in}}),
\end{equation}
where ${\bf W}_{\ell}\in\mathbb{R}^{n_{\ell}\times n_{\ell-1}}$ is the weight matrix of the $\ell$th hidden layer, $\sigma_{\ell}$ denotes the activation function, $\circ$ signifies composition, and $\bm{\theta}$ denotes all trainable parameters. Common activation functions include the hyperbolic tangent (Tanh), the rectified linear unit (ReLU), and the Gaussian error linear unit (GELU). For flow map learning, we set $n_0 = n_{L+1} = n$, where $n$ denotes the number of state variables (recalling that $\mathbf{u} \in \mathbb{R}^n$). The numbers of neurons in the hidden layers, $n_{\ell}$ with $\ell=1,2,...,L$, are hyperparameters that typically require calibration based on the specific problems.

\begin{figure}[!tbhp]
    \centering
    \includegraphics[width=0.16125\textwidth, valign=c]{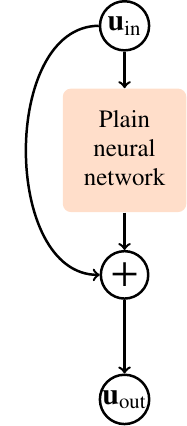}\hfill
	\includegraphics[width=0.21\textwidth, valign=c]{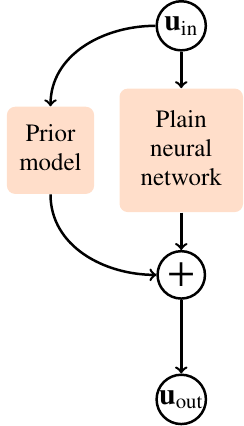}\hfill
    \includegraphics[width=0.29\textwidth, valign=c]{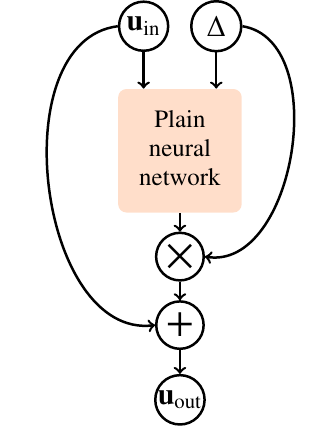}
    \caption{ResNet (left), gResNet (middle), and OSG-Net (right). The symbol ``$+$'' indicates element-wise summation, while the symbol ``$\times$'' rerpesents multiplication.}
    \label{fig:fnn}
\end{figure}

\textbf{ResNet.}
ResNet \cite{he2016deep} is an advanced variant of FNN, particularly effective for learning unknown equations \cite{qin2019data}. Initially proposed for image processing \cite{he2016deep}, ResNet introduces an identity mapping, enabling the network to learn the residue of the input-output mapping more effectively. As depicted in \Cref{fig:fnn}, a ResNet can be described as
\begin{equation}\label{eq:ResNet} 
	\widehat{\textbf{u}}_{\text{out}} = \text{ResNet}_{\bm \theta} (\textbf{u}_{\text{in}}) := \textbf{u}_{\text{in}} + {\mathcal N}_{\bm{\theta}}(\textbf{u}_{\text{in}}) = \left(  {\bf I}_n + {\mathcal N}_{\bm{\theta}} \right) ( \textbf{u}_{\text{in}} ),
\end{equation}
By comparing \eqref{eq:ResNet} with \eqref{eq:ode_fmo}, ResNet is particularly suitable for FML, as it enforces ${\mathcal N}_{\bm{\theta}}$ to approximate the effective increment of the state variables:
\begin{equation}\label{eq:resnet_increment}
	{\mathcal N}_{\bm{\theta}}(\textbf{u}_{\text{in}}) \approx \int_{0}^{\Delta} \textbf{f} (\textbf{u}(s)){\rm d}s = \int_{0}^{\Delta} \textbf{f} (\Phi_s(\textbf{u}_{\text{in}})){\rm d}s.
\end{equation}

\textbf{gResNet.} 
As shown in \Cref{fig:fnn}, gResNet \cite{chen2021generalized} generalizes the traditional ResNet concept by defining the residue as the difference between the output data and the predictions made by a \emph{prior model}:
\begin{equation}\label{eq:gResNet}
	\textbf{u}_{\text{out}}=\text{gResNet}(\textbf{u}_{\text{in}}) := \mathcal{A}(\textbf{u}_{\text{in}}) + {\mathcal N}_{\bm{\theta}} (\textbf{u}_{\text{in}}),
\end{equation}
where $\mathcal{A}$ is the prior model, and ${\mathcal N}_{\bm{\theta}}$ acts as a correction for $\mathcal{A}$. If an existing prior model is unavailable, $\mathcal{A}$ can be constructed from data, such as using a modified DMD \cite{chen2021generalized} to construct a best-fit affine model:
\begin{equation*}
	\mathcal{A}(\textbf{u}_{\text{in}}):=\textbf{A}\textbf{u}_{\text{in}}+\textbf{b},
\end{equation*}
where $\textbf{A}\in\mathbb{R}^{n\times n}$ and $\textbf{b}\in\mathbb{R}^{n}$ are determined by solving the following linear regression problem:
\begin{equation}\label{eq:mDMD}
	(\textbf{A}, \textbf{b}) = \argmin_{\substack{\tilde{\textbf{A}}\in\mathbb{R}^{n\times n}\\\tilde{\textbf{b}}\in\mathbb{R}^{n}}}\frac{1}{J}\sum_{j=1}^J\left\|\textbf{u}^{(j)}_{\text{out}} - \tilde{\textbf{A}}\textbf{u}^{(j)}_{\text{in}} - \tilde{\textbf{b}}\right\|_2^2.
\end{equation}
To solve problem (\ref{eq:mDMD}), we first augment the input vector by appending a constant term:
$$\tilde{\textbf{u}}_{\text{in}} = \begin{bmatrix}\textbf{u}_{\text{in}}\\ 1\end{bmatrix}\in\mathbb{R}^{n+1},$$
where the constant 1 accommodates the bias term $\textbf{b}$ in the affine model. Next, we construct the following matrices using the dataset (\ref{eq:data_fixed}):
$$\textbf{Y}:=[\textbf{u}^{(1)}_{\text{out}}, \textbf{u}^{(2)}_{\text{out}},\dots,\textbf{u}^{(J)}_{\text{out}}]\in\mathbb{R}^{n\times J},\quad \textbf{X}:=[\tilde{\textbf{u}}^{(1)}_{\text{in}}, \tilde{\textbf{u}}^{(2)}_{\text{in}},\dots,\tilde{\textbf{u}}^{(J)}_{\text{in}}]\in\mathbb{R}^{(n+1)\times J}.$$
The solution to the linear regression problem (\ref{eq:mDMD}) can then be explicitly expressed as
\begin{equation*}
	\begin{bmatrix}\textbf{A} & \textbf{b}\end{bmatrix} = \textbf{Y}\textbf{X}^\top(\textbf{X}\textbf{X}^\top)^{-1}.
\end{equation*}

This modification to DMD accommodates potential non-homogeneous terms in the unknown equations, making the approximation more flexible. The concept of gResNet encompasses the standard ResNet with $\textbf{A}=\textbf{I}_n$ and $\textbf{b}=\textbf{0}$.

\textbf{OSG-Net.} 
To adeptly approximate a family of flow maps associated with varying time step sizes, it is necessary to incorporate the time step size as an input to DNN. The flow maps of autonomous systems form a one-parameter semigroup, satisfying
\begin{subequations}\label{eq:semigroup_property}
	\begin{align}
		& {\Phi}_{0} = {\bf I}_n, \label{eq:semigroup_property1}\\
		\smallskip
		& {\Phi}_{\Delta_1+\Delta_2} = {\Phi}_{\Delta_1} \circ {\Phi}_{\Delta_2} \quad \forall \Delta_1,\Delta_2\in \mathbb{R}^+. 
		\label{eq:semigroup_property2}
	\end{align}
\end{subequations}
The semigroup property is crucial as it connects the system’s evolutionary behaviors across different time scales. Therefore, it is natural for data-driven models to adhere to this property. The OSG-Net, proposed in \cite{chen2023deep}, is well-suited for this purpose. Mathematically, an OSG-Net can be expressed as
\begin{equation}\label{eq:OSG-Net}
	\widehat{\textbf{u}}_{\text{out}} = \text{OSG-Net}_{\bm \theta} (\textbf{u}_{\text{in}}, \Delta) := \textbf{u}_{\text{in}} + \Delta {\mathcal N}_{\bm{\theta}}(\textbf{u}_{\text{in}}, \Delta).
\end{equation}
The architecture of OSG-Net, illustrated in \Cref{fig:fnn}, involves concatenating the state variables $\textbf{u}_{\text{in}}$ with the time step size $\Delta$ before inputting them into the network ${\mathcal N}_{\bm{\theta}}$. Unlike ResNet, OSG-Net introduces an additional skip connection that scales the output of ${\mathcal N}_{\bm{\theta}}$ by $\Delta$. This design ensures that an OSG-Net inherently satisfies the first property (\ref{eq:semigroup_property1}). As for the second property, we can design special loss functions to embed this prior knowledge into OSG-Net via training, which can enhance the model's long-term stability (see Section \ref{sec:gdsg} for detailed discussions). 

By comparing \eqref{eq:OSG-Net} with \eqref{eq:ode_fmo}, it is clear that ${\mathcal N}_{\bm{\theta}}$ serves as an approximation to the time-averaged effective increment:
\begin{equation}\label{eq:osgnet_increment}
	{\mathcal N}_{\bm{\theta}}(\textbf{u}_{\text{in}}, \Delta) \approx \frac{1}{\Delta}\int_{0}^{\Delta} \textbf{f} (\textbf{u}(s)){\rm d}s = \frac{1}{\Delta}\int_{0}^{\Delta} \textbf{f} (\Phi_s(\textbf{u}_{\text{in}})){\rm d}s.
\end{equation}
\subsection{Model Training and Prediction}
\label{sec:train_pred}

Once the data pairs are rearranged and an appropriate DNN architecture is selected, model training is carried out by minimizing a suitable loss function. The commonly used \emph{mean squared error} (MSE) quantifies the discrepancy between the predicted outputs and the actual values:
\begin{equation}\label{eq:mse}
	L(\bm{\theta})=\frac{1}{J}\sum_{j=1}^J \left\|\widehat{\textbf{u}}_{\text{out}}^{(j)}(\bm{\theta}) - \textbf{u}_{\text{out}}^{(j)}\right \|_2^2.
\end{equation}
It is worth noting that training data extracted from the same trajectory are not independent. To account for the structure of observational noise or the highly clustered nature of data from a single trajectory, a suitably weighted norm can be applied in the loss function \eqref{eq:mse}.
Some alternative loss functions  will be discussed in \Cref{sec:long-time} to enhance the prediction accuracy and stability.  

In practice, $L(\bm{\theta})$ is minimized using stochastic gradient descent (SGD) \cite{ruder2016overview} or its variants, such as Adam \cite{kingma2014adam}.  
SGD works by randomly splitting the training dataset into mini-batches. At each iteration, the gradient of the loss function with respect to $\bm{\theta}$ is computed for one mini-batch, and this gradient is used to update the parameters. This process repeats for multiple epochs until the loss function is sufficiently minimized. 
The procedure for training DNNs using SGD is outlined in \Cref{alg:sgd}.

\floatname{algorithm}{Algorithm}
\begin{algorithm}
\caption{Model training using stochastic gradient descent (SGD)}
\label{alg:sgd}
\begin{algorithmic}[1]
\Require Number of epochs $E$, batch size $B$; training data $\{(\textbf{u}_{\text{in}}^{(j)}, \textbf{u}_{\text{out}}^{(j)})\}_{j=1}^{J}$ (fixed time lag) or $\{(\textbf{u}_{\text{in}}^{(j)}, \Delta^{(j)}, \textbf{u}_{\text{out}}^{(j)})\}_{j=1}^{J}$ (varied time lags)
\State Initialize the DNN parameters $\bm{\theta}$ randomly
\For{$\text{epoch} = 1$ to $E$}
    \State Shuffle the training data
    \For{$\text{batch} = 1$ to $\left\lfloor \frac{J}{B} \right\rfloor$}
        \State Sample a mini-batch $\Lambda$ of size $B$ from the training data
        \State Update the DNN parameters:
        \[
        \bm{\theta} \leftarrow \bm{\theta} - \eta \nabla_{\bm{\theta}} L^{(\Lambda)}(\bm{\theta}),
        \]
        \State where the learning rate $\eta>0$ is often adapted during training, and 
        $$\nabla_{\bm{\theta}} L^{(\Lambda)}(\bm{\theta}) = \frac{1}{B} \sum_{j \in \Lambda} \nabla_{\bm{\theta}} \left\|\widehat{\textbf{u}}_{\text{out}}^{(j)}(\bm{\theta}) - \textbf{u}_{\text{out}}^{(j)} \right\|_2^2.$$
    \EndFor
\EndFor
\end{algorithmic}
\end{algorithm}
\floatname{algorithm}{Procedure}

Once the DNN is successfully trained, it is recursively used to conduct predictions from any given initial state 
$\textbf{u}^{\text{pre}}(t_0)=\textbf{u}(t_0)$. The trained DNN model, denoted as $\widehat{\Phi}_{\bm{\theta}}$ predicts the solution evolution as follows:
\begin{equation}\label{eq:resnet_pred}
	\textbf{u}^{\text{pre}}( t_{k+1} ) =\widehat{\Phi}_{\bm{\theta}}(\textbf{u}^{\text{pre}}( t_{k} )), \qquad k=0,1,\dots
\end{equation}
with a fixed time step size $t_{k+1}-t_k\equiv \Delta$, or
\begin{equation}\label{eq:osgnet_pred}
	\textbf{u}^{\text{pre}}( t_{k+1} ) =\widehat{\Phi}_{\bm{\theta}}(\textbf{u}^{\text{pre}}( t_k ), \Delta_k), \qquad k=0,1,\dots 
\end{equation}
with varying time step sizes $t_{k+1}-t_{k}=\Delta_k$.

\subsection{Learning Partially Observed Systems}
In many real-world scenarios, collecting data for all state variables $\textbf{u}\in \mathbb{R}^n$ is not always feasible. Instead, observations can be restricted to a subset of the state variables $\textbf{w}\in \mathbb{R}^m$, where $m<n$. This limitation shifts the focus to learning the dynamics of $\textbf{w}$ alone, resulting in \emph{non-autonomous} unknown governing equations due to the absence of other variables. Similar to the fully observed case, 
the training data can be constructed from	sampling on multiple long trajectories or many short trajectories with $M+1$ observations of $\bf w$. 
	If data from multiple long trajectories of $\textbf{w}$ with a fixed time lag $\Delta$ are available:
\begin{equation}\label{eq:partial_data}
	\textbf{w}_k^{(i)} = \textbf{w}( t_k^{(i)}; \textbf{w}_{0}^{(i)}, t_0) + {\bm \epsilon}_{\textbf{w},k}^{(i)}, \qquad k=1,\dots,K_i, \quad i=1,\dots,{I_{traj}}, 
\end{equation}
then we rearrange these trajectories into shorter bursts of $M+1$ consecutive states:
\begin{equation}\label{eq:data_partial}
    \left\{\textbf{w}_0^{(j)}, \textbf{w}_1^{(j)},\dots,\textbf{w}_{M+1}^{(j)}\right\},\qquad j=1,2,\dots,J.
\end{equation}
To model the temporal evolution of $\textbf{w}$, a memory-based DNN architecture was introduced in \cite{fu2020learning}:
\begin{equation}\label{eq:fml_partial}
    \textbf{w}_{k+1}=\textbf{w}_k + \mathcal{N}_{\bm{\theta}}(\textbf{w}_k,\textbf{w}_{k-1},\dots,\textbf{w}_{k-M}),\qquad k\geq M>0,
\end{equation}
where $T:=M\Delta$ represents the memory length, which is problem-dependent and often requires manual tuning. The state $\textbf{w}_k$ at time $t_k$, along with the $M$ preceding states, are concatenated as inputs for the neural network $\mathcal{N}_{\bm{\theta}}$.
The following loss function is then minimized:
\begin{equation}\label{eq:loss_memory}
    L(\bm{\theta})=\frac{1}{J}\sum_{j=1}^J \left\|\textbf{w}_{M+1}^{(j)}-\left(\textbf{w}_M^{(j)}+\mathcal{N}_{\bm{\theta}}(\textbf{w}_M^{(j)},\dots,\textbf{w}_1^{(j)},\textbf{w}_0^{(j)})\right)\right \|_2^2.
\end{equation}
Learning a fully observed system is a special case with $m=n$ and $M=0$. Once the DNN model is successfully trained, it can be recursively used to predict the system's evolution from any initial states  $\left(\textbf{w}(t_0),\textbf{w}(t_1),\dots,\textbf{w}(t_M)\right)$:
\begin{equation}\label{eq:partial_pred}
	\resizebox{0.9\hsize}{!}{ $ \displaystyle \begin{cases}
		\textbf{w}^{\text{pre}}(t_k)=\textbf{w}(t_k),\qquad k=0,1,\dots,M,\\
	    \textbf{w}^{\text{pre}}( t_{k+1} ) = \textbf{w}^{\text{pre}}(t_k) + \mathcal{N}_{\bm{\theta}}\left(\textbf{w}^{\text{pre}}(t_k),\textbf{w}^{\text{pre}}(t_{k-1}),\dots,\textbf{w}^{\text{pre}}(t_{k-M})\right), \quad k\geq M,
	\end{cases} $}
\end{equation}
where $t_{k+1}-t_k\equiv \Delta$. 

This approach has also been applied to systems with hidden parameters \cite{fu2022modeling}, as well as PDE systems with snapshot data observed on a subset of the domain \cite{churchill2023dnn}.
\subsection{Learning Unknown PDEs}
\label{sec:pde_modal}
\begin{figure}[tbhp]
    \centering
    \includegraphics[width=0.8\linewidth]{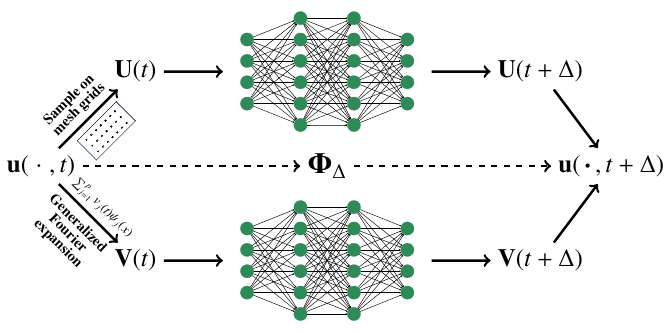}
    \caption{Learning PDEs in nodal space (top branch) and modal space (bottom branch).}
    \label{fig:nodal_modal}
\end{figure}

The aforementioned framework can be seamlessly extended to data-driven modeling of unknown PDEs. 
This can be effectively achieved in either nodal or modal space, as illustrated in \Cref{fig:nodal_modal}.

\subsubsection{Learning in Nodal Space} 
Let ${\bf u}:\Omega\times\mathbb{R}^+\rightarrow\mathbb{R}^{d_u}$ represent the state variables of the underlying unknown $d$-dimensional PDE, and $\Omega\subset\mathbb{R}^d$, where $d$ is the spatial dimension, and $d_u$ is the length of the state vector $\bf u$. As shown in the upper branch of \cref{fig:nodal_modal}, assume we have measurement data of $\bf u$ at a set of nodal points ${\mathbb X} = \left\{ x_1, \dots, x_n \right\} \subset \Omega $, collected from various trajectories:
\begin{equation}\label{eq:pde_data_nodal}
    \textbf{U}_k^{(i)} = \textbf{U}(t_k^{(i)};\textbf{U}_0^{(i)}, t_0) + {\bm \epsilon}_{\textbf{U},k}^{(i)},\qquad k=1,\dots,K_i,  \qquad i=1,\cdots, I_{traj},
\end{equation} 
where $\textbf{U}(t)=\left({\bf u}(x_1,t), \dots, {\bf u}(x_n,t) \right)^\top\in\mathbb{R}^{n\times d_u}$ is a matrix. 
While ResNet and OSG-Net built upon FNNs can be used for learning PDEs \cite{chen2022deep,chen2023deep}, they can be computationally expensive when $\mathbb X$ contains a large number of nodal points. 
To address this, we can replace FNNs with more suitable DNNs, such as the convolutional neural networks (CNNs) \cite{lee2019data,xu2020multi}, 
the Fourier Neural Operator (FNO) \cite{li2021fourier}, and many other neural operators \cite{li2020neural,cao2021choose,chen2024positional}, including those built upon Transformers \cite{vaswani2017attention,chen2024positional} from large language models.

{\bf Transformers.} Transformers \cite{vaswani2017attention}, particularly those based on the self-attention mechanism, are highly effective for capturing long-range dependencies in data. Mathematically, a generalized Transformer can be expressed as
\begin{equation*}
	\mathcal{T}_{\bm{\theta}}(\textbf{U}_{\text{in}}) = {\omega}_{L+1} \circ ( \sigma_L \circ  \alpha_L \circ \omega_L ) \circ \cdots \circ ( \sigma_1 \circ \alpha_1 \circ \omega_1 ) (\textbf{U}_{\text{in}}, \mathbb{X}),
\end{equation*}
where each set of operations $\{\sigma_\ell \circ \alpha_\ell \circ \omega_\ell\}_{\ell=1}^L$ represents the following transformation:
\begin{equation}\label{eq:posatt}
	\textbf{U}_\ell = \sigma_{\ell} \circ \alpha_\ell \circ \omega_\ell (\textbf{U}_{\ell-1}):= \sigma_{\ell}(A_\ell\textbf{U}_{\ell-1}W_\ell).
\end{equation}
Here, $\textbf{U}_\ell \in \mathbb{R}^{n_\ell \times d_\ell}$ is a matrix, with $\ell=1,2,\dots,L$, represents the output of the \( \ell \)-th hidden layer. The initial input, $\textbf{U}_0 = [\textbf{U}_{\text{in}}, \mathbb{X}] \in \mathbb{R}^{n \times (d_u + d)}$, is formed by concatenating the input function values and nodal point coordinates. In this setup: $\sigma_{\ell}$ is the activation function; $\omega_\ell$ represents a transformation via right multiplication by a weight matrix $W_\ell \in \mathbb{R}^{d_{\ell-1} \times d_\ell}$; $\alpha_\ell$ represents a convolution via left multiplication by a kernel matrix $A_\ell \in \mathbb{R}^{n_{\ell} \times n_{\ell-1}}$. 
Each hidden layer can thus be interpreted as transforming a vector-function with $d_{\ell-1}$ components sampled on a latent grid $\mathbb{X}_{\ell-1} = \{x_{\ell, j}\}_{j=1}^{n_{\ell-1}}$, to a new vector-function with $d_{\ell}$ components sampled on a new latent grid $\mathbb{X}_{\ell} = \{x_{\ell, i}\}_{i=1}^{n_{\ell}}$, where $\mathbb{X}_0 = \mathbb{X}_L = \mathbb{X}$. 
The sizes of the hidden layers, specified by $\{d_{\ell}\}_{\ell=1}^L$ and $\{n_{\ell}\}_{\ell=1}^{L-1}$, are hyperparameters that typically require tuning based on the problem at hand. At the output layer, we set $d_{L+1} = d_u$ and $n_L = n$ to produce the predicted function values on the target grid $\mathbb{X}$.

Transformers can be enhanced with a \emph{multi-head attention mechanism}, performing multiple convolutions in each hidden layer to provide a comprehensive view of the target operator. This is achieved by replacing $A_\ell\textbf{U}_{\ell-1}W_\ell$ in \eqref{eq:posatt} with the concatenation of different heads $\{A^h_\ell\textbf{U}_{\ell-1}W^h_\ell\}_{h=1}^H$, where $A^h_\ell\in\mathbb{R}^{n_{\ell}\times n_{\ell-1}}$ and $ W^h_\ell\in\mathbb{R}^{d_{\ell-1}\times \frac{d_\ell}{H}}$. 

The general formulation in \eqref{eq:posatt} encompasses many deep learning methods, distinguished by the implementation of the convolution operator $A_{\ell}$.
\begin{itemize}
	\item {\bf In CNNs} \cite{lecun1995convolutional}, $A_{\ell}$ performs local weighted sums over spatially structured data. The non-zero values of $A_{\ell}$, which constitute the trainable weights, are identical but shifted accross the rows, as these weights are shared accross $\Omega$. 
This convolution is usually combined with pooling or up-pooling layers \cite{ronneberger2015u}, which downsample or upsample $\textbf{U}_\ell$ from the grid $\mathbb{X}_{\ell-1}$ to a coarser or finer grid $\mathbb{X}_{\ell}$.

    \item {\bf In Transformers} built upon the self-attention mechanism \cite{vaswani2017attention}, $A_{\ell}$ performs global convolution. Mathematically, $A_{\ell}$ is implemented as
	\begin{equation}\label{eq:selfatt}
		A_{\ell} = \text{Softmax}\left(\frac{(\textbf{U}_{\ell-1}W^Q_{\ell})(\textbf{U}_{\ell-1}W^K_{\ell})^\top}{\sqrt{d_{\ell-1}}}\right),
	\end{equation}
	where $W^Q_{\ell},W^K_{\ell}\in\mathbb{R}^{d_{\ell-1}\times d_{\ell}}$ are two trainable weight matrices, and Softmax normalizes each row of a matrix into a discrete probability distribution. In  \cite{li2022transformer}, a cross-attention mechanism was proposed to enable the change of mesh. Specifically, $\textbf{U}_{\ell-1}W_{\ell}^Q$ in \eqref{eq:selfatt} is replaced by $\mathbb{X}_{\ell}W^X_{\ell}$, with $W^X_{\ell}\in\mathbb{R}^{d\times d_{\ell}}$ being a trainable weight matrix. This design allows cross-attention to output a new function sampled on any mesh $\mathbb{X}_{\ell}$. 
\end{itemize}

{\bf Position-induced Transformer (PiT).} 
Here, we present a Transformer-based method, named PiT, built upon the position-attention mechanism proposed in \cite{chen2024positional}. 
Distinguished from other Transformer-based networks  \cite{cao2021choose,hao2023gnot,li2022transformer} built upon the classical self-attention \cite{vaswani2017attention}, position-attention implements the convolution operator by considering the spatial interrelations between sampling points. Define the pariwise distance matrix $D_{\ell}\in\mathbb{R}^{n_{\ell}\times n_{\ell-1}}$ between $\mathbb{X}_{\ell}$ and $\mathbb{X}_{\ell-1}$ by  
$D_{\ell,ij}=\|x_{\ell,i}-x_{\ell-1,j}\|^2_2.$  
Then $A_{\ell}$ is defined as
$
    A_{\ell}:=\text{Softmax}(-\lambda_{\ell} D_{\ell}), 
$
where $\lambda_{\ell}\in\mathbb{R}^+$ is a trainable parameter. 
Position-attention represents a global linear convolution with a stronger focus on neighboring regions, resonating with the concept of \emph{domain of dependence} in PDEs and making PiT appealing for learning PDEs \cite{chen2024positional}. The parameter $\lambda_{\ell}$ is interpretable, as most attention at a point  $x_{\ell,i}\in\mathbb{X}_{\ell}$ is directed towards those points $x_{\ell-1,j}\in\mathbb{X}_{\ell-1}$ with the distance to $x_{\ell,i}$ smaller than $1/\sqrt{\lambda_{\ell}}$. In practice, we construct a latent mesh $\mathbb{X}_{\text{ltt}}$ by coarsening $\mathbb{X}$ while preserving essential geometric characteristics, and let
$\mathbb{X}_{\ell}=\mathbb{X}_{\text{ltt}},\quad n_{\ell}=n_{\text{ltt}},\quad \text{for}\;\;\ell=1,2,\dots,L-1,$
with $n_{\text{ltt}}<n$. This design reduces the computational cost caused by a potential large number of sampling points in the dataset. 
Like many other neural operators \cite{kovachki2023neural,azizzadenesheli2024neural}, PiT is mesh-invariant and discretization convergent. Once trained, PiT can generalize to new input meshes, delivering consistent and convergent predictions as the input mesh is refined. 
To learn time-dependent unknown PDEs from data, we construct a (g)ResNet or OSG-Net with PiT as the basic block.  
Once the model is successfully trained, we can recursively call the model to predict the evolutionary behaviors of ${\bf u}(x,t)$ given any initial conditions.

\subsubsection{Learning in Modal Space}\label{sec:modal_pde}
An alternative strategy is to model unknown PDEs in modal space \cite{wu2020data} by combining traditional model reduction with deep learning approaches. Initially, select a finite-dimensional function space with a suitable basis to approximate each component of ${\bf u}(x,\cdot)$: 
\begin{equation*}
	\mathbb V^p = \text{span}\left\{ \psi_1(x),...,\psi_p(x)\right\}, 
\end{equation*}
where $p\leq n$, and the basis functions ${\bf \Psi}(x) := \left( \psi_1(x),...,\psi_p(x) \right)^\top$ are defined on the physical domain $\Omega$. As shown in the lower branch of \cref{fig:nodal_modal}, the solution of the underlying PDE can then be approximated in $\mathbb V^p$ by a finite-term series:
\begin{equation*}
	{\bf u}(x,t) \approx \sum_{j=1}^p {\bf v}_j(t)\psi_j(x),
\end{equation*}
with $\textbf{V}:=\left( {\bf v}_1,..., {\bf v}_p \right)^\top \in \mathbb{R}^{p \times d_u}$ being the modal expansion coefficients. This introduces a bijective mapping:
\begin{equation}\label{eq:bijective}
 \Pi:\mathbb{R}^{p \times d_u}\to [\mathbb V^p]^{d_u}, \qquad   \Pi\textbf{V} =  {\bf V}^\top {\bf \Psi}(x),
\end{equation}
which defines a unique correspondence between a function in $[\mathbb V^p]^{d_u}$ and its modal expansion coefficients.

Now, we project each data sample 
	$
	\textbf{U}_k^{(i)}
	$ in \eqref{eq:pde_data_nodal}  
	 into $[\mathbb V^p]^{d_u}$, yielding a coefficient matrix $\mathbf{V}_k^{(i)}$. This is achieved by solving the linear regression problem:
\begin{equation}\label{eq:leastsq}
	\textbf{V}_k^{(i)}=\argmin_{\tilde{\textbf{V}} \in\mathbb{R}^{p \times d_u}} \left\| ( \textbf{U}_k^{(i)} )^\top -  \tilde{\textbf{V}}^\top {\bf \Psi}(\mathbb{X}) \right\|_2^2,
\end{equation}
where ${\bf \Psi}(\mathbb{X}) = \left({\bf \Psi}(x_1), {\bf \Psi}(x_2),\dots,{\bf \Psi}(x_n)\right)$ is a $p\times n$ matrix, representing the basis function values evaluated at the sampling grids $\mathbb X$. The solution to \cref{eq:leastsq} can be expressed as
\begin{align*} 
	\textbf{V}_k^{(i)}&=\left({\bf \Psi}(\mathbb{X}){\bf \Psi}(\mathbb{X})^\top\right)^{-1}{\bf \Psi}(\mathbb{X})\textbf{U}_k^{(i)}.
\\
    & = \textbf{V}(t_k^{(i)};\textbf{V}_0^{(i)}, t_0) + {\bm \epsilon}_{\textbf{V},k}^{(i)},\qquad k=1,\dots,K_i,  \qquad i=1,\cdots, I_{traj},
\end{align*}
where $\textbf{V}(t_k^{(i)};\textbf{V}_0^{(i)}, t_0)$ denotes the modal coefficients of the underlying function, and ${\bm \epsilon}_{\textbf{V},k}^{(i)}=\left({\bf \Psi}(\mathbb{X}){\bf \Psi}(\mathbb{X})^\top\right)^{-1}{\bf \Psi}(\mathbb{X}){\bm \epsilon}_{\textbf{U},k}^{(i)}$ represents the noise inherited from the nodal value noise. 
We can then treat $\textbf{V}$ as the state variables and model the unknown governing ODEs using deep learning approaches, offering a predictive model for the evolution of $\textbf{V}$. The behavior of $\textbf{U}$ can be easily inferred through the bijective mapping (\ref{eq:bijective}).

Learning unknown PDEs in the modal space provides great flexibility in choosing different basis functions to represent the solution, including trigonometric functions, wavelet functions, Legendre polynomials, and piecewise polynomials. This approach is analogous to traditional numerical methods, such as spectral Galerkin, finite element, and finite volume methods, commonly used for solving known PDEs.

\subsubsection{Remarks on Learning PDEs} 
In the modal learning approach, when an interpolation basis is used, the resulting modal coefficients directly correspond to function values. This allows both the modal and nodal learning approaches to be represented through the expansion shown in the bottom path of \Cref{fig:nodal_modal}, highlighting a connection between the two methods. Although Transformers were originally developed for nodal learning, they may also be adapted for modal learning, as the attention mechanism can be used to capture dependencies among different modes.

Our data-driven models in DUE serve as approximate evolution operators for the underlying unknown PDEs. They enable prediction of future solutions for any initial conditions without necessitating retraining. This contrasts with physics-informed neural networks (PINNs) \cite{raissi2019physics}, which require fewer or no measurement data but solve a given PDE for a specific initial condition, typically necessitating retraining for each new initial condition.

The above deep learning frameworks are not only useful for modeling unknown PDEs but also for creating surrogate models of known, yet complex, PDEs that are expensive to solve using traditional numerical methods.

\section{Enhancing Prediction Accuracy and Stability}
\label{sec:long-time}
In learning unknown time-dependent differential equations, our goal is to predict the system's evolution accurately over extended periods. This section introduces two loss functions and a novel neural network architecture designed to enhance the long-term prediction accuracy and stability of the learned models.

\subsection{Multi-step Loss}
Research by \cite{chen2022deep} shows that using a multi-step loss function can significantly improve predictive models with fixed time step sizes. This approach averages the loss over multiple future time steps. The training dataset is structured as follows:
\begin{equation}\label{eq:data_multistep}
	\left\{\textbf{w}^{(j)}_0, \textbf{w}^{(j)}_1,\dots,\textbf{w}^{(j)}_{M+1},\dots,\textbf{w}^{(j)}_{M+1+K}\right\},\qquad j=1,2,\dots,J,
\end{equation}
where $K\geq 0$ represents the number of future time steps. During training, initial states $\textbf{w}^{(j)}_0, \textbf{w}^{(j)}_1,\dots,\textbf{w}^{(j)}_M$ are used, and the DNN model (\ref{eq:fml_partial}) is executed for $K+1$ steps to produce predictions $\widehat{\textbf{w}}^{(j)}_{M+1},\dots,\widehat{\textbf{w}}^{(j)}_{M+1+K}$. The multi-step loss function is defined as 
\begin{equation}\label{eq:multistep_loss}
	L(\bm{\theta})=\frac{1}{J(K+1)}\sum_{j=1}^J \sum_{k=0}^K \left\|\textbf{w}^{(j)}_{M+1+k}-\widehat{\textbf{w}}^{(j)}_{M+1+k}(\bm{\theta})\right \|_2^2. 
\end{equation}
Note that the loss function in \Cref{eq:loss_memory} is a special case with $K=0$.

\subsection{Semigroup-informed Loss}
\label{sec:gdsg}
As mentioned in Section \ref{sec:dnn}, an OSG-Net inherently satisfies the first constraint (\ref{eq:semigroup_property1}). To embed the second property (\ref{eq:semigroup_property2}) into an OSG-Net, a \emph{global direct semigroup-informed} (GDSG) loss function was introduced in \cite{chen2023deep}, which effectively guides an OSG-Net to adhere to (\ref{eq:semigroup_property2}) through training. The GDSG method integrates a regularization term informed by the semigroup property (\ref{eq:semigroup_property2}) to the data-driven loss function: 
\begin{equation}\label{eq:loss_sg}
	L({\bm \theta})=\frac{1}{(1+\lambda)J}\sum_{j=1}^J \left(\left\|\textbf{u}_{\text{out}}^{(j)}-\widehat{\textbf{u}}^{(j)}_{\text{out}}(\bm{\theta})\right \|_2^2 + \lambda 
	R_{SG}^{(j)} ( {\bm \theta} ) \right),
\end{equation}
where $\lambda>0$ serves as a regularization factor, and $R_{SG}^{(j)}( {\bm \theta} )$ is defined as
\begin{equation}\label{eq:gdsg}
	R_{SG}^{(j)} ( {\bm \theta} ) := 	\frac{1}{2}\left( \left\|\bar{\textbf{u}}^{(j)} ({\bm \theta}) - \tilde{\textbf{u}}^{(j)} ({\bm \theta}) \right\|_2^2 + \left\|\bar{\textbf{u}}^{(j)} ({\bm \theta}) - \breve{\textbf{u}}^{(j)} ({\bm \theta}) \right\|_2^2 \right),
\end{equation}
with $\bar{\textbf{u}}^{(j)}, \tilde{\textbf{u}}^{(j)}$, and $\breve{\textbf{u}}^{(j)}$ being network predictions of randomly generated initial conditions $\widetilde{\textbf{u}}_0^{(j)}$ and random forward time steps $\Delta_0^{(j)}$, $\Delta_1^{(j)}$:
\begin{equation*}
	{\bar{\textbf{u}}^{(j)}}  = \text{OSG-Net}_{\bm{\theta}} \left(\widetilde{\textbf{u}}_0^{(j)}, \Delta_0^{(j)}+\Delta_1^{(j)} \right),
\end{equation*}
which is the predicted state after a single forward step of size $\Delta_0^{(j)}+\Delta_1^{(j)}$, and
\begin{align*}
	{\tilde{\textbf{u}}^{(j)}} &= \text{OSG-Net}_{\bm{\theta}} \left(\text{OSG-Net}_{\bm{\theta}} \left(\widetilde{\textbf{u}}_0^{(j)}, \Delta_0^{(j)}\right), \Delta_1^{(j)}\right),\\
	{\breve{\textbf{u}}^{(j)}} &= \text{OSG-Net}_{\bm{\theta}}\left(\text{OSG-Net}_{\bm{\theta}} \left(\widetilde{\textbf{u}}_0^{(j)}, \Delta_1^{(j)} \right), \Delta_0^{(j)} \right),
\end{align*}
which are the predicted states after two sequential forward steps. 
According to the semigroup property, $\bar{\textbf{u}}^{(j)}, \tilde{\textbf{u}}^{(j)}$, and $\breve{\textbf{u}}^{(j)}$ are predictions of the same true state and should therefore be enforced to be equal. 
 Hence, incorporating (\ref{eq:gdsg}) into the loss function encourages $\text{OSG-Net}_{\bm{\theta}}$ to adhere to property (\ref{eq:semigroup_property2}). Remarkably, computing the residue (\ref{eq:gdsg}) does not require additional measurement data. Moreover, the GDSG method can be further improved by generating multiple pairs of random data $\{\widetilde{\textbf{u}}_0^{(j,q)}, \Delta_0^{(j,q)}, \Delta_1^{(j,q)}\}_{q=1}^Q$ and using the averaged residue over $Q$ pairs; see Section 3.2 of \cite{chen2023deep} for more details.
\subsection{Dual-network Technique for Multiscale Dynamics} 

Modeling equations with varying time step sizes necessitates capturing dynamics characterized by temporal multiscale properties. A plain neural network may struggle with large time scale separations, leading to poor long-term prediction accuracy. In this paper, we introduce a novel dual-network architecture, called the dual-OSG-Net, which we propose as a new approach that leverages the gating mechanism \cite{hochreiter1997long} to effectively learn dynamics across broader time scales.

\begin{figure}[tbhp]
	\centering
	\includegraphics[width=0.75\linewidth]{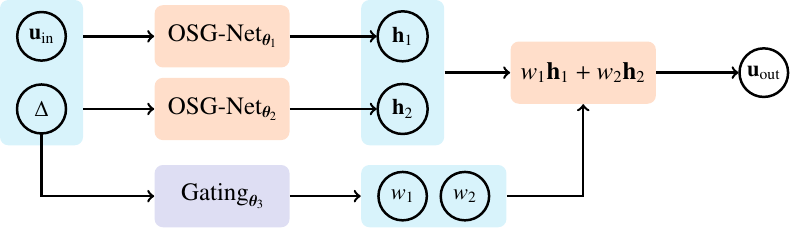}
	\caption{Dual-OSG-Net for learning multiscale equations.}
	\label{fig:dual_osgnet}
\end{figure}

As illustrated in \Cref{fig:dual_osgnet}, the dual-OSG-Net combines predictions from two independent OSG-Nets using weighted averaging. The weights $\{(w_1, w_2)|w_1>0, w_2>0, w_1+w_2=1\}$ are determined by another neural network, $\text{Gating}_{\bm{\theta}_3}$, with Softmax activation at its output layer. This gating network $\text{Gating}_{\bm{\theta}_3}$ is trained simultaneously with the two OSG-Nets ($\text{OSG-Net}_{\bm{\theta}_1}$ and $\text{OSG-Net}_{\bm{\theta}_2}$) and intelligently decides which OSG-Net weighs more. 
The gating mechanism adaptively assigns a weight to each OSG-Net based on the time step size, allowing each network to adaptively focus on a specific scale. For small time steps, it prioritizes the OSG-Net optimized for fine-scale dynamics, while for larger steps, it emphasizes the network suited to coarse scales. This adaptability enables the dual-OSG-Net to handle multi-scale problems more effectively than a single, larger OSG-Net, which lacks this flexibility and must attempt to capture all scales simultaneously. In Section \ref{sec:Robertson}, we will demonstrate the superior performance of the dual-OSG-Net compared to the standard single OSG-Net through numerical comparisons.

%% file: 3_Usage.tex
\section{Overview and Usage of DUE}
\label{sec:overview}
This section introduces the structure and usage of DUE, a comprehensive library designed for data-driven learning of unknown equations. As illustrated in \Cref{fig:due}, DUE comprises three main modules:
\begin{itemize}
    \item {\tt \textbf{datasets}}: This module handles data loading and essential preprocessing tasks such as slicing, regrouping, and normalization.
    \item {\tt \textbf{networks}}: It includes a variety of DNN architectures like FNN, ResNet, gResNet, OSG-Net, dual-OSG-Net, Transformers, and more.
    \item {\tt \textbf{models}}: This module is dedicated to training the deep learning-based models, offering various learning strategies to enhance prediction accuracy and stability.
\end{itemize}
This structure allows users to quickly understand its usage and customize or add new functionalities as needed. Detailed usage and customization of DUE are explained in Sections \ref{sec:usage} and \ref{sec:custom}.


\subsection{Usage}
\label{sec:usage}

With DUE, learning unknown equations is simplified to just a few lines of code. 
Below is a template script with detailed comments for modeling the dynamics of a damped pendulum (see Section \ref{sec:pendulum} for detailed descriptions). For more complex tasks, slight modifications may be needed, such as alternating data loaders, changing neural network architectures, and adapting training strategies.

\begin{center}
   \includegraphics[width=.99\textwidth]{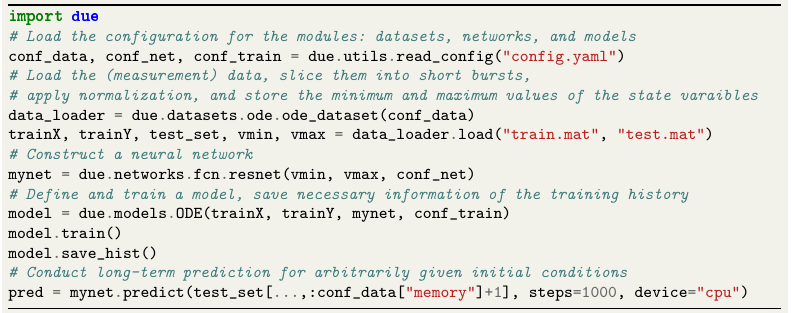}
\end{center}
\subsubsection{Configuration}
\label{sec:config}
To simplify the specification of hyperparameters such as memory steps, multi-steps in the loss function, network depth and width, training epochs, batch size, and more, users can consolidate them in a single configuration file. This file can be seamlessly processed using the {\tt \textbf{due.utils.read\_config}} function. Upon processing, these hyperparameters are stored in three Python dictionaries: one for data processing configuration, one for neural network architecture configuration, and one for model training configuration. All the modules in  \Cref{fig:due} are designed to work with such dictionaries, relieving users from specifying each hyperparameter individually when calling multiple modules. This streamlined approach facilitates the launch of new tasks and allows for easy calibration of hyperparameters. 
To automate hyperparameter optimization \cite{feurer2019hyperparameter}, users can implement automated grid search via an external script that iterates over the configuration file in a for-loop.

\begin{center}
   \includegraphics[width=.99\textwidth]{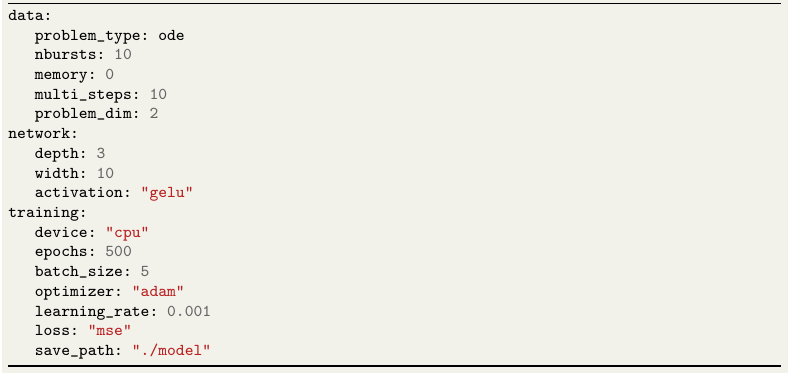}
\end{center}
\subsubsection{Data Preprocessing}

DUE is equipped to handle both ODE and PDE data with either fixed or varied time lags. To accommodate these diverse scenarios, we have implemented four modules in the ``{\tt \textbf{datasets}}'' class: 
\begin{itemize}
	\item {\tt \textbf{ode\_dataset}}: For unknown ODEs and data with fixed time lag.
	\item {\tt \textbf{ode\_dataset\_osg}}: For unknown ODEs and data with varied time lags.
	\item  {\tt \textbf{pde\_dataset}}: For unknown PDEs and data with fixed time lag.
	\item {\tt \textbf{pde\_dataset\_osg}}: For unknown PDEs and data with varied time lags.
\end{itemize}
Users only need to prepare the measurement data and employ one of these four modules. The data will be automatically rearranged, normalized, and partitioned into input-output pairs, as indicated by (\ref{eq:data_fixed}), (\ref{eq:data_varied}), (\ref{eq:data_partial}), and (\ref{eq:data_multistep}).
\subsubsection{Neural Networks}
\label{sec:nn}
The {\tt \textbf{networks}} module in DUE offers a wide array of DNN architectures for ODE and PDE learning. For modeling ODEs with fixed and varied time step sizes, we have implemented {\tt \textbf{resnet}}, {\tt \textbf{gresnet}}, and {\tt \textbf{osg\_net}} built upon FNNs, respectively. As for learning PDEs, we have implemented {\tt \textbf{pit}}---the Position-induced Transformer \cite{chen2024positional}---for handling data with fixed time lag, and {\tt \textbf{osg\_fno}}---an OSG-Net built upon the Fourier neural operator \cite{li2021fourier,chen2023deep}---for cases with varied time step sizes. As described in Section \ref{sec:pde_modal}, unknown PDEs can also be learned in modal space. We provide the {\tt \textbf{generalized\_fourier\_projection1d}} and {\tt \textbf{generalized\_fourier\_projection2d}} functions for computing modal expansion coefficients from snapshot data for one- and two-dimensional problems. All the DNN architectures in DUE belong to the {\tt \textbf{nn}} class, which can be further enriched by customized deep learning methods to suit specific needs.
\subsubsection{Model Training}
The {\tt \textbf{models}} module implements the training procedures for deep learning models. Four training routines are available:
\begin{itemize}
	\item {\tt \textbf{ode}}: For learning unknown ODEs with fixed time step size.
	\item {\tt \textbf{ode\_osg}}: For modeling unknown ODEs with varied time step sizes.
	\item  {\tt \textbf{pde}}: For learning unknown PDEs  with fixed time step size
	\item {\tt \textbf{pde\_osg}}: For modeling unknown PDEs with varied time step sizes.
\end{itemize}
We have also integrated the GDSG method to embed the semigroup property into models with varied time step sizes. Users only need to specify the hyperparameters of the semigroup loss as detailed in Section
 \ref{sec:gdsg}, and DUE handles the complex procedures of training with the GDSG method.
\subsection{Customization}
\label{sec:custom}
We have adopted a modular architecture for DUE, ensuring that its key modules, {\tt \textbf{networks}} and {\tt \textbf{models}}, can be separately customized. Users have the flexibility to adapt the neural network architecture to suit their specific requirements and implement new training methods to enhance models' prediction accuracy and stability. In this section, we briefly show how to customize neural network architectures and training methods.
\subsubsection{Neural Networks}
As described in Section \ref{sec:nn}, DUE already provides a range of neural network architectures that address various scenarios in ODE and PDE learning. Users interested in exploring more specialized or recent deep learning methods can implement them by following the guidelines in Procedure 4.1.

\begin{center}
   \includegraphics[width=.99\textwidth]{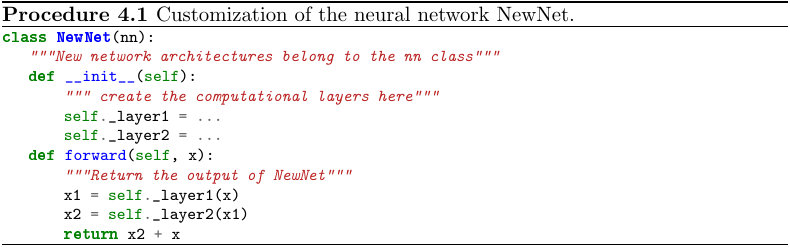}
\end{center}
\subsubsection{Model Training}
In the current version of DUE, we have implemented the multi-step loss function \cite{chen2022deep} for data-driven modeling with fixed time step size, and the GDSG method \cite{chen2023deep} for cases with varied time step sizes. If users have developed custom training methods, such as new loss functions, implementing them in DUE is straightforward using the following procedure.
\begin{center}
   \includegraphics[width=.99\textwidth]{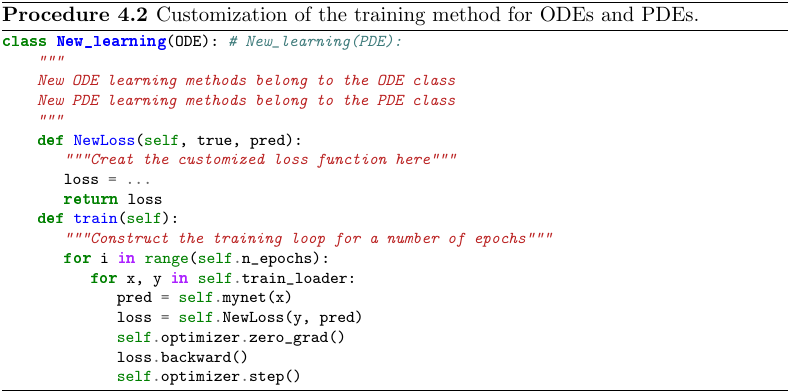}
\end{center}

By leveraging DUE’s modularity and flexibility, users can effectively address a wide range of data-driven modeling challenges in unknown ODE and PDE systems.

%% file: 4_Examples.tex
\section{Demonstration Examples}
\label{sec:experiments}

In this section, we present diverse examples to demonstrate the effectiveness of DUE for data-driven learning of unknown ODEs and PDEs. The examples include: (1) the damped pendulum system, (2) coupled oscillators with real-world noisy data, (3) the chaotic Lorenz system, (4) the Robertson chemical reaction problem involving high stiffness and multi-scale dynamics, (5) the one-dimensional viscous Burgers' equation, and (6) the vorticity evolution of the two-dimensional Navier--Stokes equations, and (7) the two-dimensional flow past a circular cylinder. In these examples, the true governing equations are known, but they serve only two purposes: generating synthetic data for training the DNNs and providing reference solutions for comparison during testing. During the data-driven learning process, the true equations are regarded as unknown.

For all examples, we use the GELU activation function \cite{hendrycks2016gaussian} and train the models with the Adam optimizer \cite{kingma2014adam}. The learning rate is initialized at 0.001 and follows a cosine annealing schedule \cite{loshchilov2016sgdr}. Detailed training configurations and dataset information for all examples are presented in the dedicated subsections below. To ensure users can quickly understand how DUE works and apply it to their tasks, we provide detailed comments in the code for each numerical example. All this information is available on the GitHub page of DUE.

\subsection{Damped Pendulum}
\label{sec:pendulum}
The first example is the damped pendulum system \cite{qin2019data,churchill2023flow}, described by the equations
\eqref{dampedPS} 
with $\alpha=0.1$ and $\beta=9.80665$. Synthetic data is generated using the fourth-order Runge--Kutta method to advance the true system forward in time. The dataset comprises $N=1,000$ trajectories for $(u_1,u_2)$, each with a length of $L=1,000$ and a time lag of $\Delta=0.02$. The initial states of these trajectories are randomly sampled from the uniform distribution on $\Omega=[-\pi/2, \pi/2]\times [-\pi, \pi]$.

\subsubsection{Fully Observed Case}
\label{sec:pendu_full}
\begin{figure}[th!]
    \centering
        \includegraphics[width=0.335\linewidth, valign=t]{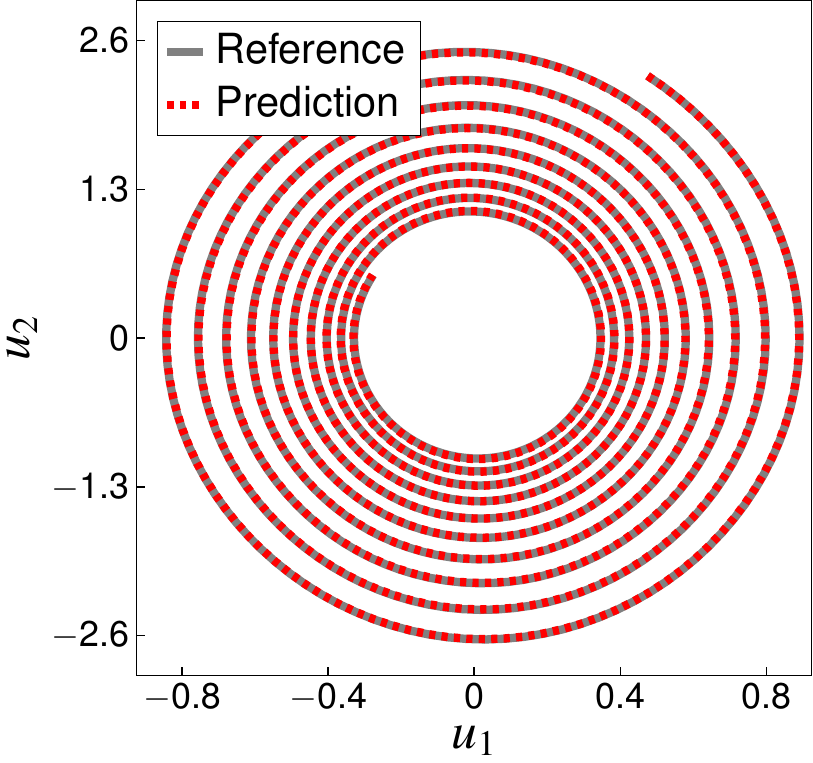}
        \includegraphics[width=0.64\linewidth, valign=t]{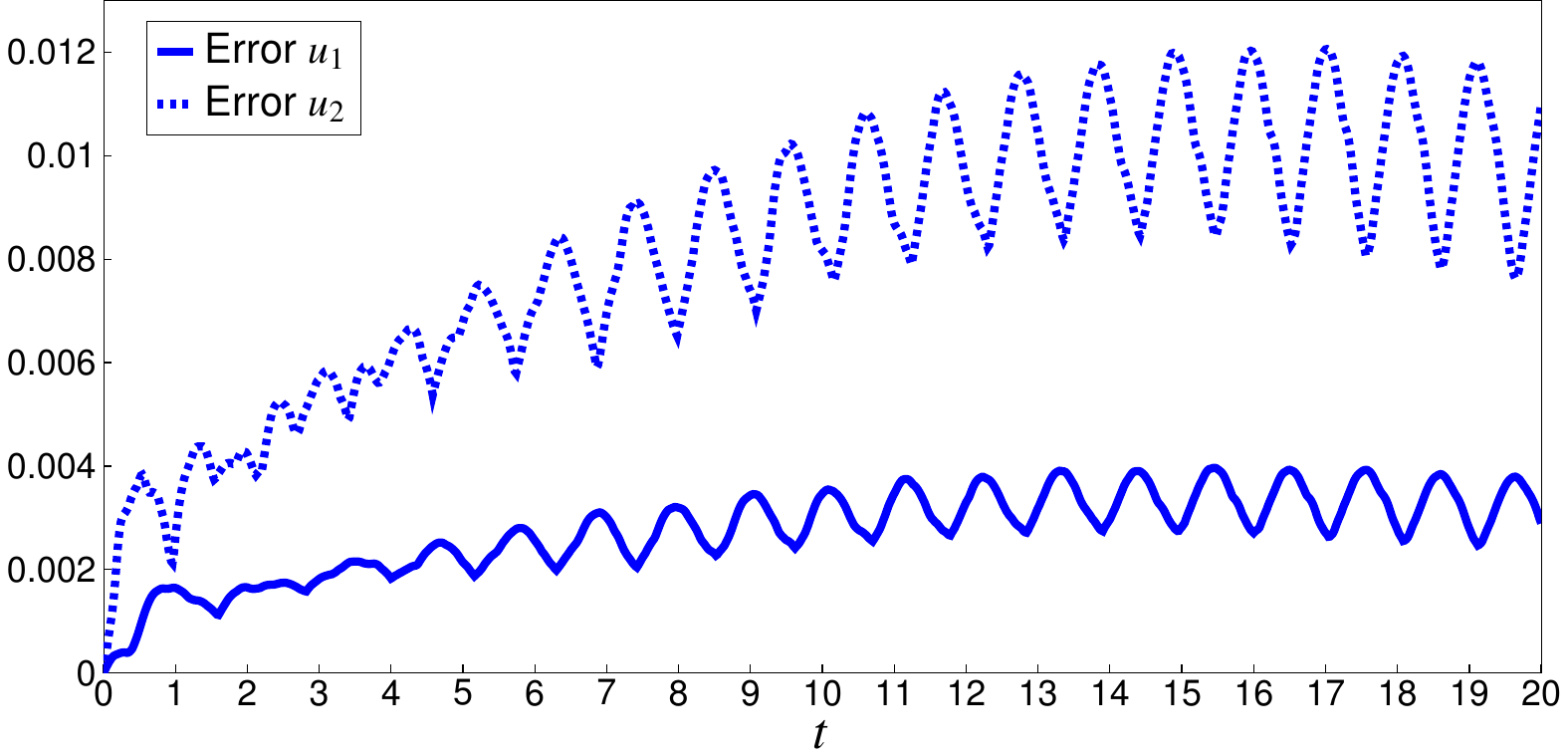}
    \caption{Fully observed damped pendulum system. Left: Comparison between the predicted and reference solutions. Right: Average $\ell_2$ error computed on the test set.}
    \label{fig:pendu_full}%
\end{figure}

In this case, we assume both state variables $u_1$ and $u_2$ are observable. We set $K=10$ for the multi-step loss and randomly sample 10 bursts from each trajectory to construct the training dataset. The ResNet has 3 hidden layers, each with 10 neurons, and is trained for 500 epochs with a batch size of 5. Following training, the model's performance is evaluated on a new and unseen test set consisting of 100 trajectories with initial states uniformly sampled on $\Omega$. \Cref{fig:pendu_full} displays an example trajectory alongside the reference solution, as well as the average $\ell_2$ error over time. The trained model demonstrates accurate predictions up to $t=20$, equivalent to 1,000 forward steps.

\subsubsection{Partially Observed Case}
\label{sec:pendu_partial}
\begin{figure}[th!]
    \centering
        \includegraphics[width=0.48\linewidth, valign=t]{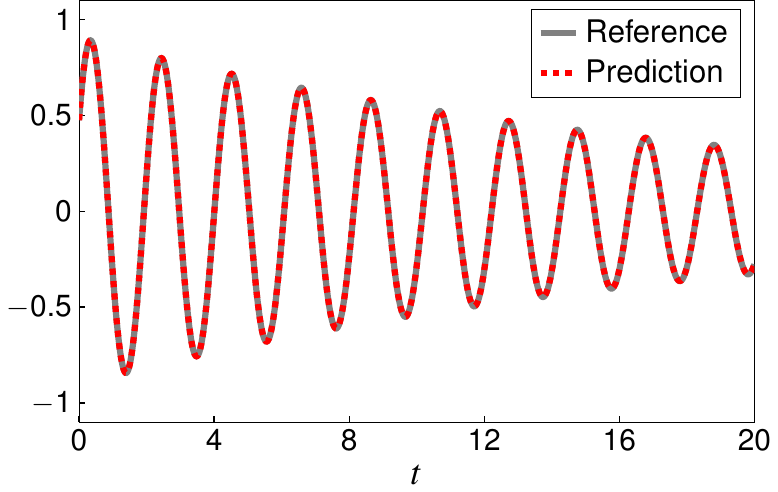}
        \includegraphics[width=0.49\linewidth, valign=t]{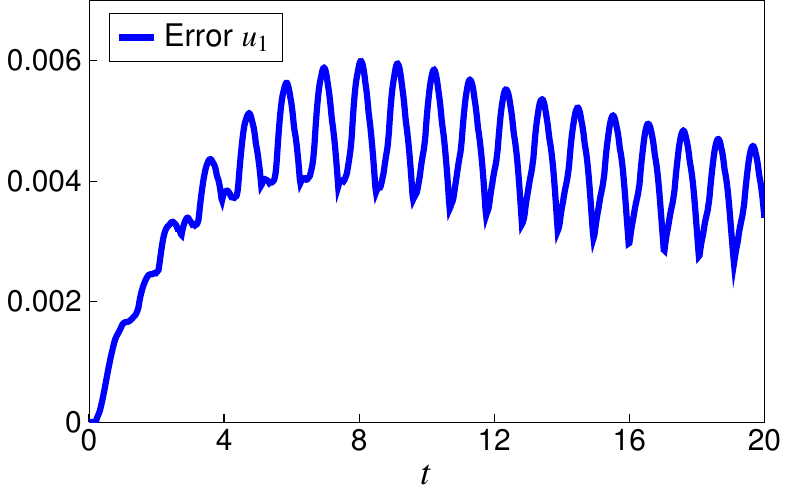}
    \caption{Partially observed damped pendulum system. Left: Comparison between the predicted and reference solutions. Right: Average $\ell_2$ error computed on the test set.}
    \label{fig:pendu_partial}
\end{figure}

In this scenario, we focus on modeling a reduced system solely related to $u_1$. Trajectories of $u_2$ are excluded from the training data, and we address this partially observed system by adopting $M=10$ memory steps in the model. Thanks to the optimized data processing module of DUE, users can easily try different values of $M$ by modifying the {\tt \textbf{memory}} parameter in the configuration file; see Section \ref{sec:config}. Other configurations remain the same as in the fully observed case. \Cref{fig:pendu_partial} illustrates an example trajectory and the average $\ell_2$ error on the test set.

\subsubsection{Robustness to Noisy Data}
\begin{figure}[th!]
    \centering
        \includegraphics[width=0.48\linewidth, valign=t]{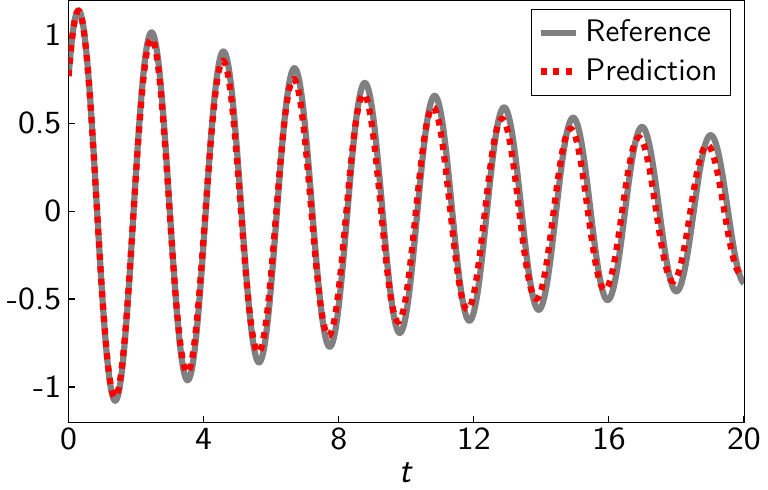}
        \includegraphics[width=0.49\linewidth, valign=t]{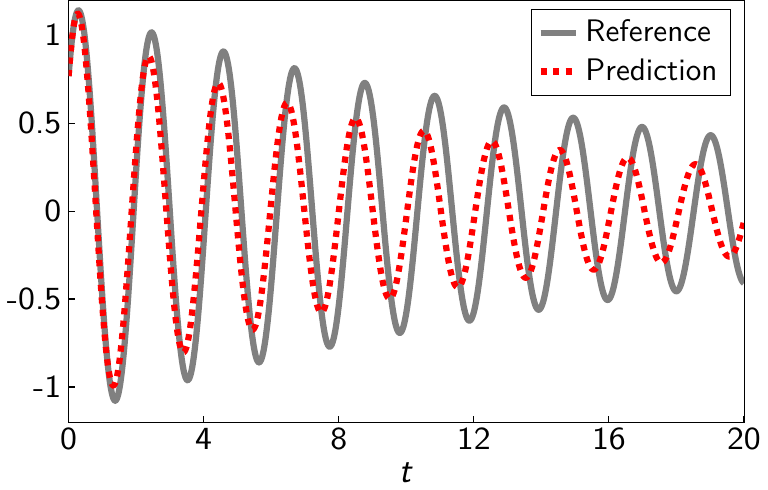}
    \caption{Partially observed damped pendulum system with noisy data. Left: Noise level $\eta=5\%$. Right: Noise level $\eta=10\%$.}
    \label{fig:pendu_noise}
\end{figure}

In the third scenario, we introduce artificial noise to the synthetic data used in Section \ref{sec:pendu_partial} to assess the model's robustness to measurement errors. Specifically, the training data are modified as
\begin{equation*}
	\left\{\textbf{u}_{\text{in}}^{(j)}(1+{\bm \epsilon}_{\text{in}}^{(j)}), \textbf{u}_{\text{out}}^{(j)}(1+{\bm \epsilon}_{\text{out}}^{(j)})\right\}_{j=1}^{J},
\end{equation*}
where the relative noise terms ${\bm \epsilon}_{\text{in}}^{(j)}$ and ${\bm \epsilon}_{\text{out}}^{(j)}$ are drawn from a uniform distribution over $[-\eta, \eta]$, with $\eta$ representing the noise level. We perform two experiments with $\eta$ set to $0.05$ and $0.1$, corresponding to noise levels of $5\%$ and $10\%$, respectively. All other settings are kept the same as in Section \ref{sec:pendu_partial}. 
\Cref{fig:pendu_noise} shows the predicted trajectories generated by two different models trained on noisy data. While some deviation from the exact dynamics is observed, the oscillating and damping patterns of the solution remain well-captured. The model's performance can be further enhanced by increasing the amount of training data.

\subsection{Two Coupled Oscillators with Real-World Noisy Data}
\label{sec:pendu_double}

\begin{figure}[thb!]
    \centering
    \includegraphics{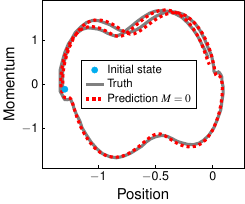}\hfill
    \includegraphics{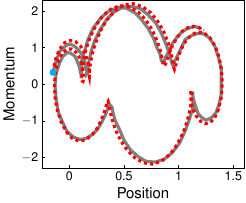}\hfill
    \includegraphics{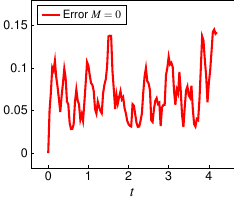}\vspace{0.2cm}
    \includegraphics{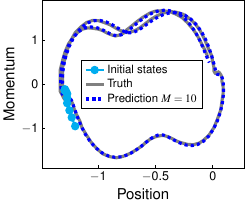}\hfill
    \includegraphics{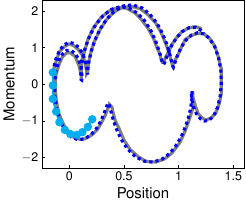}\hfill
    \includegraphics{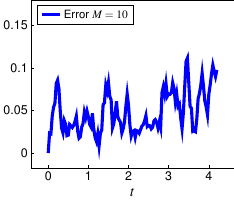}
    \caption{Two coupled oscillators from real-world data. Models with different memory steps ($M$) predict the system's evolution. Top: $M=0$. Bottom: $M=10$. Left: Phase plots of Mass 1. Middle: Phase plots of Mass 2.  Right: the $\ell_{\infty}$ error (suggested in \cite{zhu2023implementation}) computed on the last 126 states of the experimental data.}
    \label{fig:pendu_double}
\end{figure}

Next, we use DUE to model the unknown ODEs of two coupled oscillators using real-world data \cite{schmidt2009distilling,zhu2023implementation}. This dataset consists of a single trajectory with $486$ recorded states, of which the first $360$ states are used for training and the remaining for testing. The state variables of interest include the positions and momenta of the two oscillators, resulting in a state space in  $\mathbb{R}^4$. Due to measurement noise, the experimental data may not perfectly represent the full system. In this example, we examine the impact of memory terms in modeling partially observed systems by training two models with $M=0$ and $M=10$, respectively. Each model employs a ResNet with 3 hidden layers, each containing 10 neurons, and is trained for 500 epochs with a batch size of 1. The predicted phase plots are displayed in \Cref{fig:pendu_double}. Despite the data scarcity and measurement noise, both models successfully capture the underlying dynamics. The advantage of using memory terms is evident from the improved accuracy with $M=10$ compared to $M=0$.

\subsection{Chaotic Lorenz system}
\label{sec:lorenz}
\begin{figure}[thb!]
    \centering
    \includegraphics[width=0.32\linewidth]{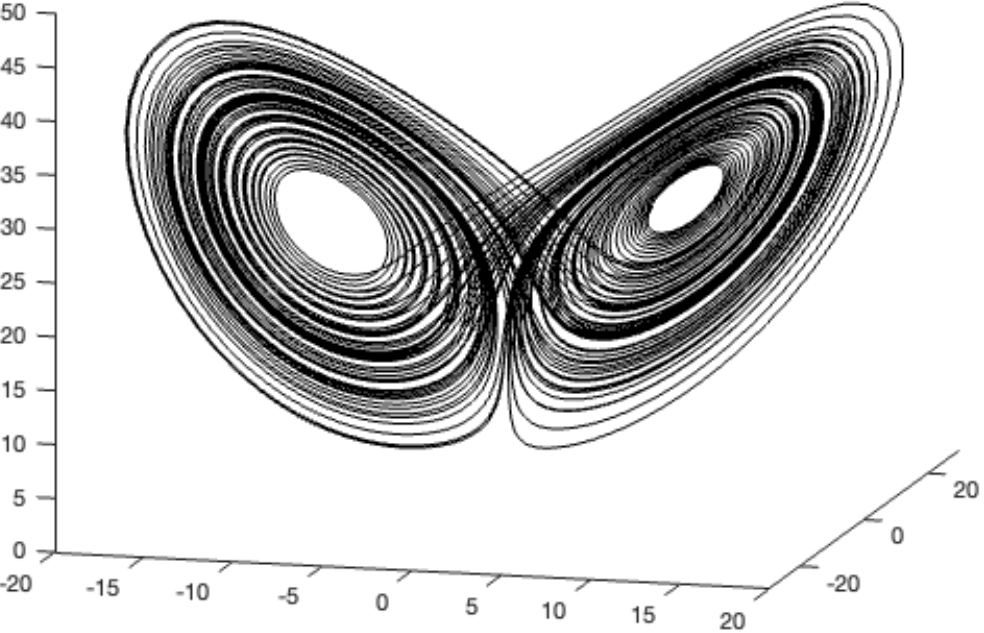}\hfill
    \includegraphics[width=0.32\linewidth]{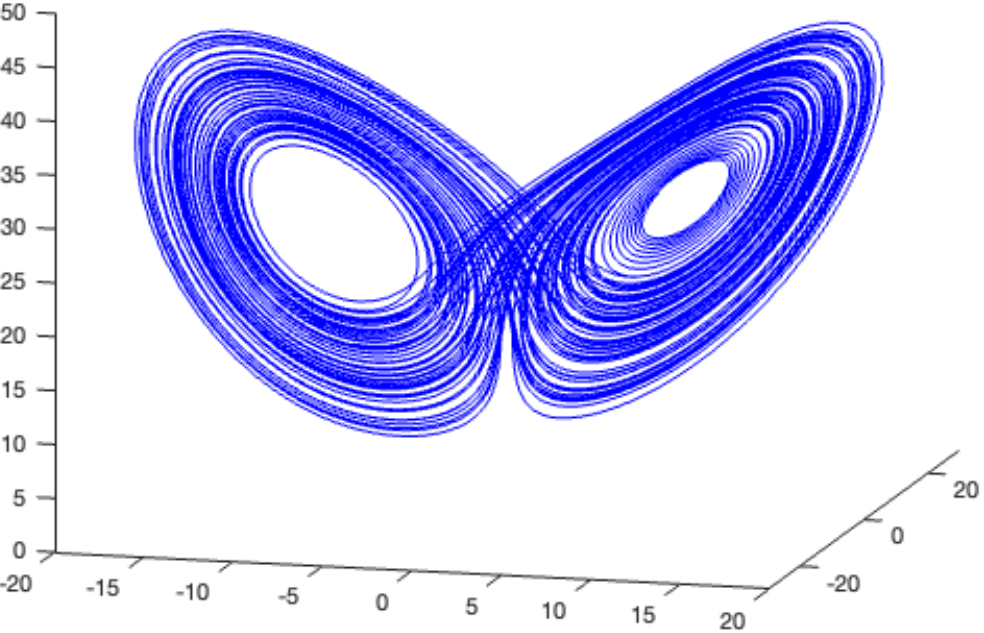}\hfill
    \includegraphics[width=0.32\linewidth]{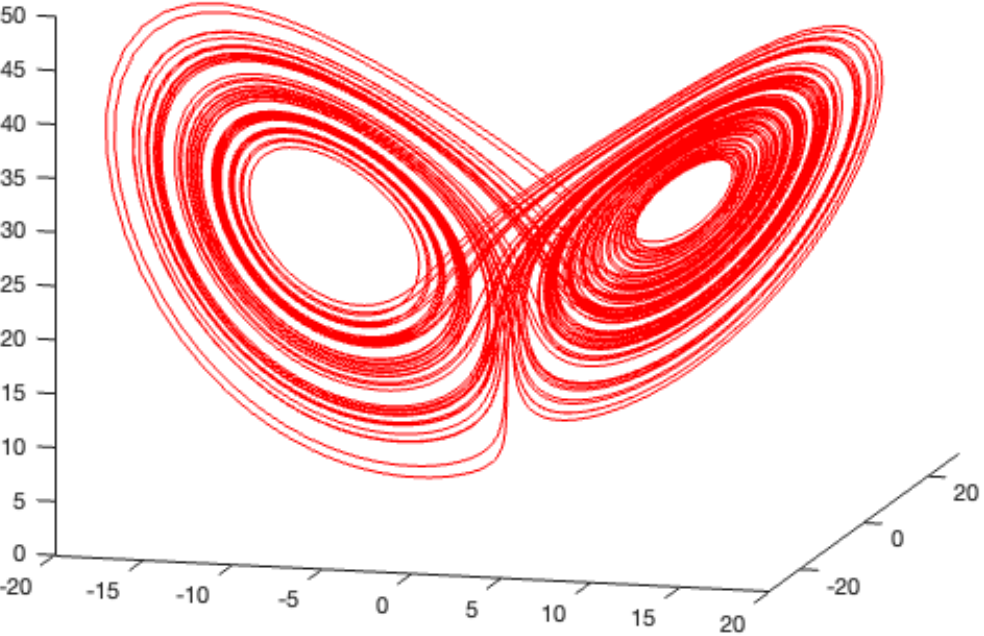}
    \caption{Lorenz equations. From left to right: reference solution, prediction by gResNet, prediction by ResNet.}
    \label{fig:lorenz_phase}
\end{figure}

\begin{figure}[thb!]
    \centering
    \includegraphics[width=0.333\linewidth]{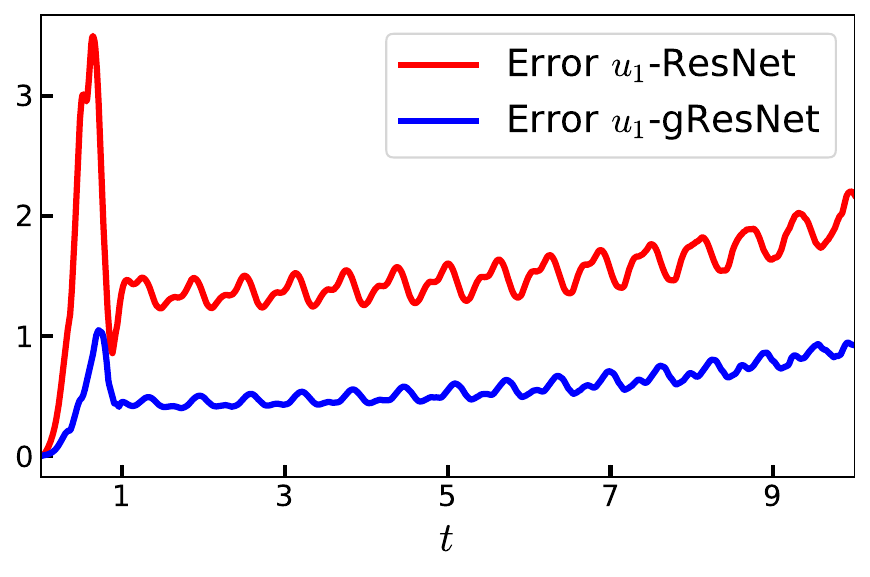}\hfill
    \includegraphics[width=0.333\linewidth]{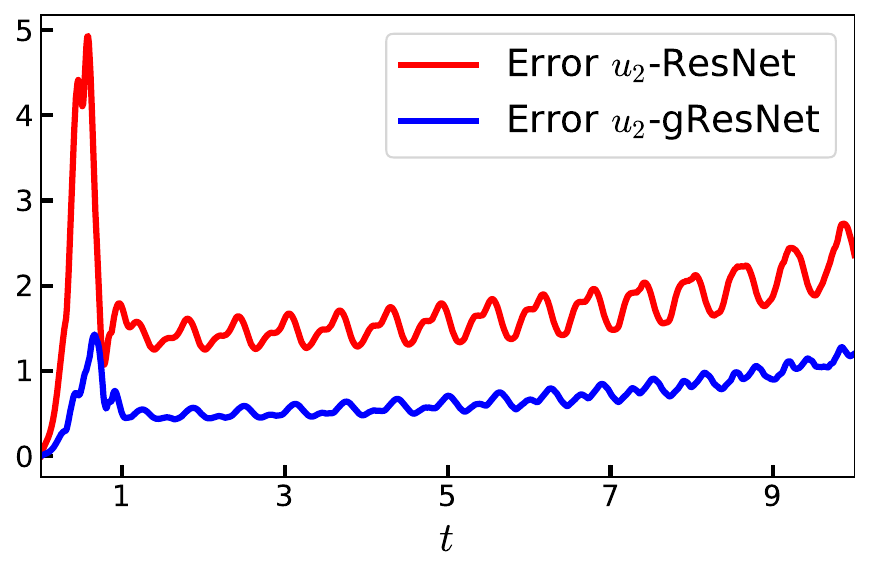}\hfill
    \includegraphics[width=0.333\linewidth]{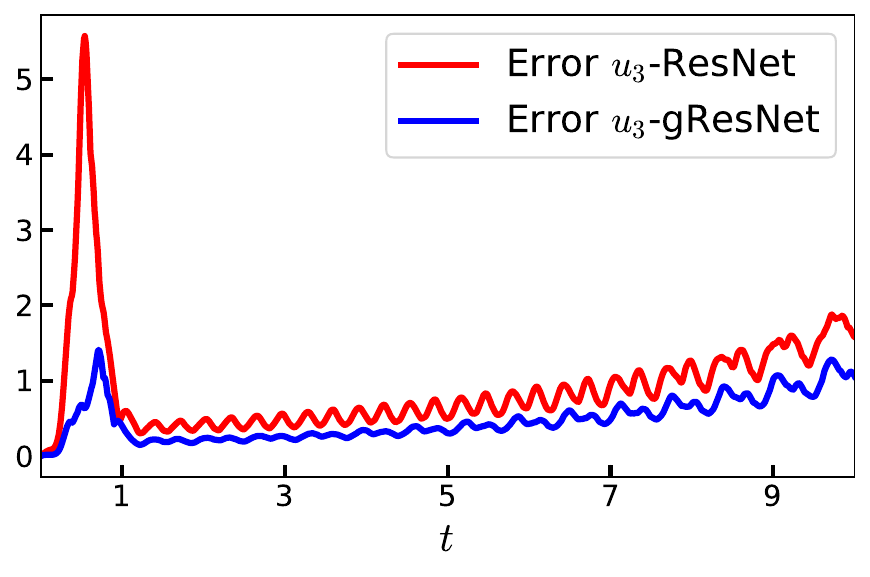}
    \caption{Lorenz equations. From left to right: $\ell_2$ error of $u_1$, $u_2$, and $u_3$.}
    \label{fig:lorenz_error}
\end{figure}

Next, we demonstrate DUE's capability to model the chaotic Lorenz system \cite{churchill2022deep}. The true  equations are given by:
\begin{equation}
	\begin{dcases}
		\frac{d u_1}{dt}=\sigma(u_2-u_1),\\
		\frac{d u_2}{dt}=u_1(\rho-u_3) - u_2,\\
		\frac{d u_3}{dt}=u_1u_2-\beta u_3,
	\end{dcases}
	\label{eq:lorenz}
\end{equation}
with $\sigma=10$, $\rho=28$, and $\beta=8/3$. The synthetic dataset consists of $N=1,000$ trajectories for $(u_1, u_2, u_3)$, each with a length of $L=10,000$ and a time lag of $\Delta=0.01$. Initial states are randomly sampled from the uniform distribution on $\Omega=[-\pi/2, \pi/2]^3$. 
We set $K = 10$ for the multi-step loss and randomly sample 5 bursts from each trajectory to construct the training dataset. In this example, we compare the performance of ResNet and gResNet. The baseline ResNet is built upon an FNN with 3 hidden layers, each with 10 neurons. The gResNet consists of an FNN with the same architecture and a pre-trained affine model, implemented as {\tt \textbf{affine}} in DUE. Both models are trained for 500 epochs with a batch size of 5. 
After training, the models are evaluated on a new and unseen test set consisting of 100 trajectories with initial states uniformly sampled on $\Omega$. \Cref{fig:lorenz_phase} displays an example of the predicted and reference trajectories, while \Cref{fig:lorenz_error} shows the average $\ell_2$ error up to $t=10$ on the test set. These results indicate that both ResNet and gResNet can capture the system's chaotic evolution, with gResNet achieving higher prediction accuracy.

\subsection{Robertson Chemical Reaction Equations with Multi-Scale Dynamics}
\label{sec:Robertson}

This example explores the Robertson chemical reaction system, which describes the kinetics of three chemical species: A, B, and C. Proposed by Robertson in 1966 \cite{robertson1966solution}, the system is governed by the following nonlinear ODEs:
\begin{equation}
	\begin{dcases}
		\frac{d u_1}{dt}=-k_1u_1 + k_2u_2u_3,\\
		\frac{d u_2}{dt}=k_1u_1 - k_2u_2u_3 - k_3u_2^2,\\
		\frac{d u_3}{dt}=k_3u_2^2,
	\end{dcases}
	\label{eq:Robertson}
\end{equation}
where $(u_1,u_2,u_3)$ represent the concentrations of $(A,B,C)$, respectively. The reaction rates are $k_1=0.04$, $k_2=10^4$, and $k_3=3\times 10^7$, making the system highly \emph{stiff}. To capture dynamics across both small and large time scales, we use DUE's {\tt \textbf{ode\_osg}} module to approximate flow maps with varied time step sizes \cite{chen2023deep}. 
The synthetic dataset comprises $50,000$ input-output pairs, with time lags randomly sampled from  $10^{U[-4.5,2.5]}$, where $U[-4.5,2.5]$ is the uniform distribution on $[-4.5,2.5]$. Initial states are randomly sampled from the domain $[0,1]\times [0,5\times 10^{-5}]\times [0,1]$, and the system is solved using the variable-step, variable-order {\tt ode15s} solver in \textsc{Matlab}.

\begin{figure}[thb!]
    \centering
    \includegraphics[width=0.48\linewidth]{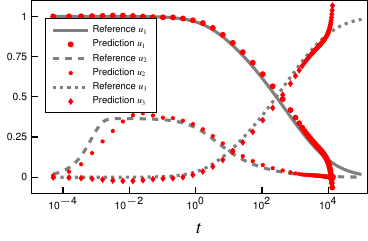}
    \includegraphics[width=0.48\linewidth]{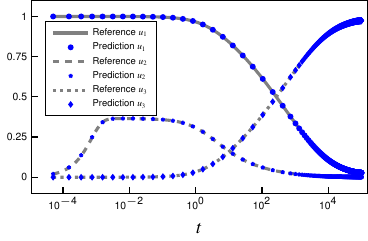}
\caption{Robertson chemical reaction equations. Left: OSG-Net prediction vs.~reference solution. Right: Dual-OSG-Net prediction vs.~reference solution. Initial state: $(1,0,0)$. The value of $u_2$ is multiplied by $10^4$ for clearer visualization.}
    \label{fig:Robertson}
\end{figure}

To address the challenge of multi-scale temporal dynamics, we employ a dual-OSG-Net with 3 hidden layers, each containing 60 neurons. The neural network model is trained using the GDSG method to embed the semigroup property, with the hyperparameters $\lambda$ and $Q$ both set to 1. Additionally, we train a second model using the vanilla OSG-Net \cite{chen2023deep} for benchmarking. Both models are trained for 10,000 epochs with a batch size of 500. 
After training, predictions are initiated from $(u_1,u_2,u_3)=(1,0,0)$ to forecast the multi-scale kinetics of the three chemical species until $t=100,000$, a challenging long-term prediction task. The time step size starts from $\Delta_1=5\times 10^{-5}$ and doubles after each step until it reaches $\Delta_2=300$. As shown in \Cref{fig:Robertson}, the dual-OSG-Net model accurately predicts the dynamics across all time scales between $\Delta_1$ and $\Delta_2$, demonstrating superior long-term accuracy compared to the vanilla OSG-Net model.

\subsection{One-dimensional Viscous Burgers' Equation}
\label{sec:Burgers}

This example demonstrates DUE's capabilities in learning PDEs by focusing on the viscous Burgers' equation with Dirichlet boundary conditions \cite{wu2020data,chen2022deep}:
\begin{equation}
	\label{eq:Burgers}
	\begin{dcases}
		\partial_t u +  \partial_x \left(\frac{u^2}{2}\right)= \frac{1}{10} \partial_{xx}u, \quad & (x,t)\in (0,2\pi)\times \mathbb{R}^+,\\
		u(0,t) = u(2\pi,t)  = 0, & t \ge 0.
	\end{dcases}
\end{equation}
The training data are generated by sampling the power series solutions of the true equation on a uniform grid with 128 nodal points. Initial conditions are drawn from a Fourier series with random coefficients: 
$u(x,t=0)=\sum_{m=1}^{10} a_m\sin (mx)$,  
where \( a_m\sim U[-1/m,1/m] \). We generate \( N=1,000 \) trajectories of the solution with different initial conditions, and record \( L=40 \) snapshots on each trajectory with a time lag \( \Delta=0.05 \).

\begin{figure}[th!]
    \centering
    \includegraphics[width=0.49\textwidth]{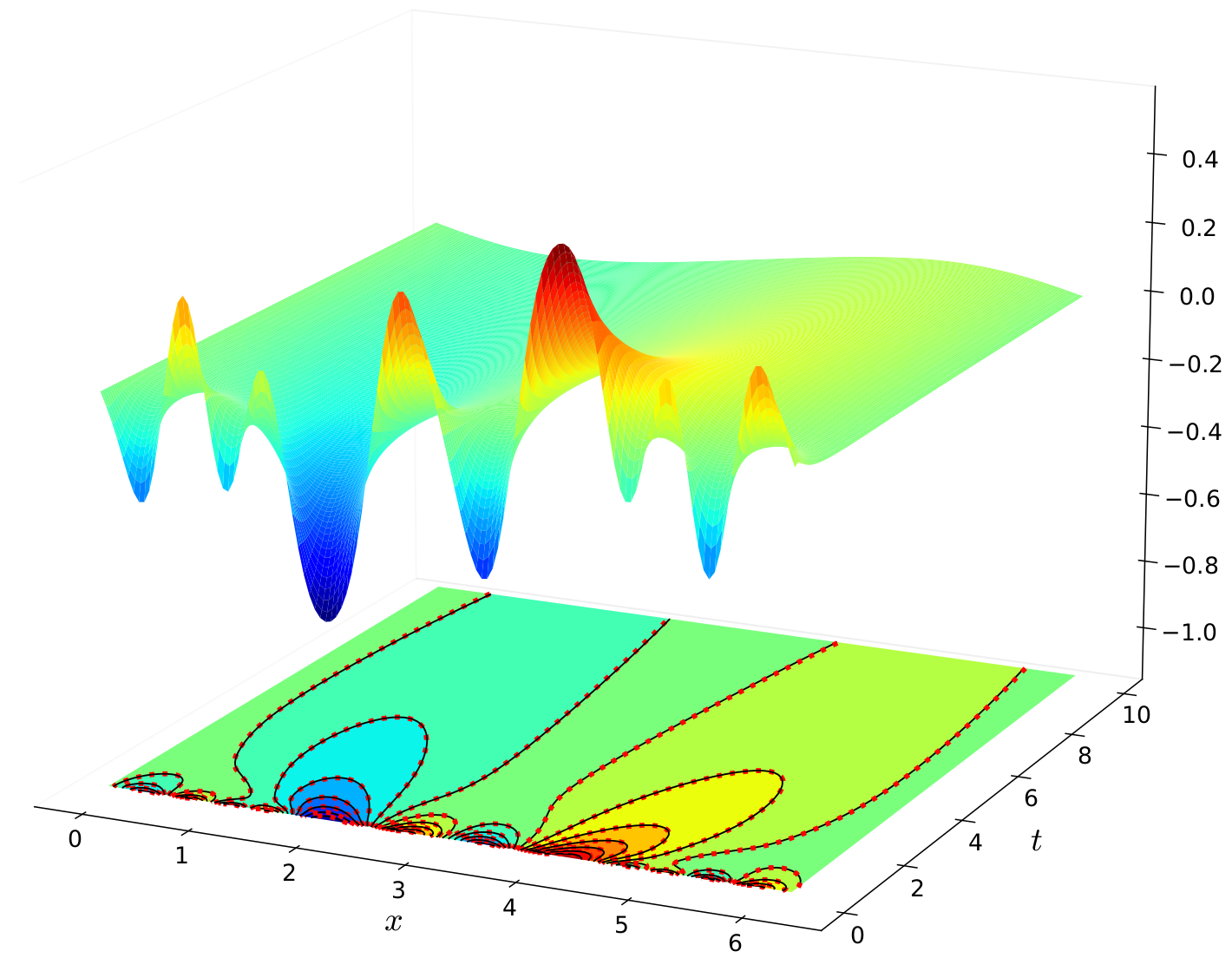}
    \includegraphics[width=0.49\textwidth]{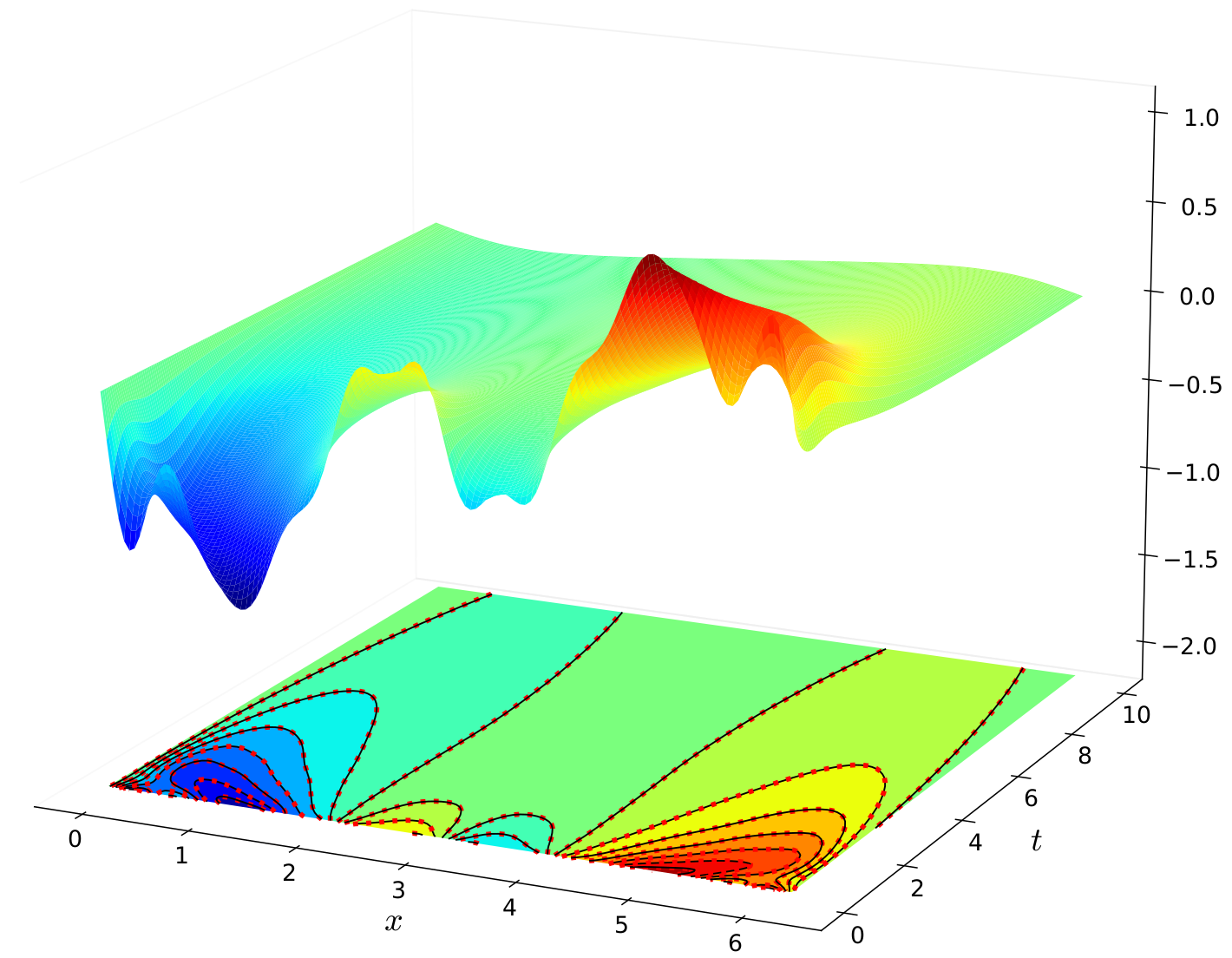}
\caption{One-dimensional viscous Burgers' equation. Predicted and reference solutions for two example trajectories originating from different initial conditions. Black solid lines indicate reference solution contours, while red dotted lines and colored plots show predictions.}
    \label{fig:Burgers}
\end{figure}

In this example, we introduce how to learn PDEs in modal space using DUE's {\tt \textbf{generalized\_fourier\_projection1d}} class. First, initialize this class by specifying a truncation wave number for the modal expansion. The training data are projected into the reduced modal space via the {\tt \textbf{generalized\_fourier\_projection1d.forward}} function. This data transformation is followed by a standard ODE modeling procedure, resulting in a model that captures the dynamics of the modal coefficients. During prediction, the future states of the modal coefficients are used to recover solutions in the physical space using the {\tt \textbf{generalized\_fourier\_projection1d.backward}} function. 
For this example, the truncation wave number is set to 10. We adopt a ResNet with 3 hidden layers, each containing 60 neurons. The model is trained for 500 epochs with a batch size of 10. After training, we evaluate the model's performance on a new and unseen test set. \Cref{fig:Burgers} displays predictions for two example trajectories up to \( t=10 \), equivalent to 200 forward steps.

\subsection{Two-dimensional Incompressible Navier--Stokes Equations}
\label{sec:ns}
\begin{figure}[thb!]
\centering
        \includegraphics[width=0.2\textwidth, valign=c]{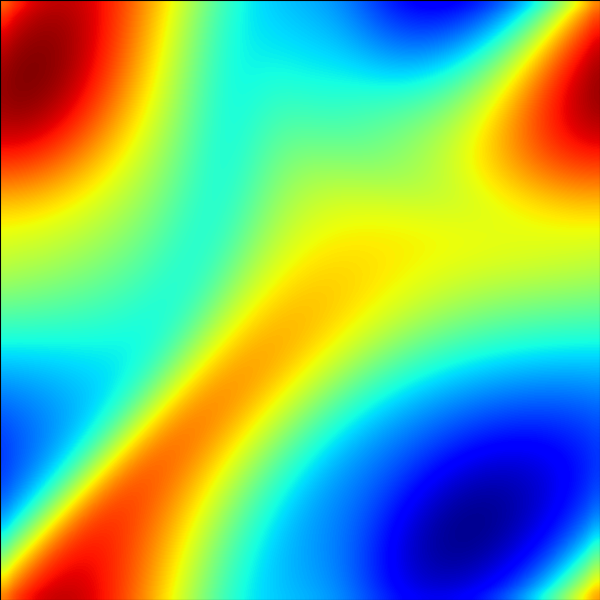}
        \includegraphics[valign=c]{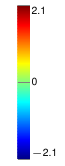}
        \includegraphics[width=0.2\textwidth, valign=c]{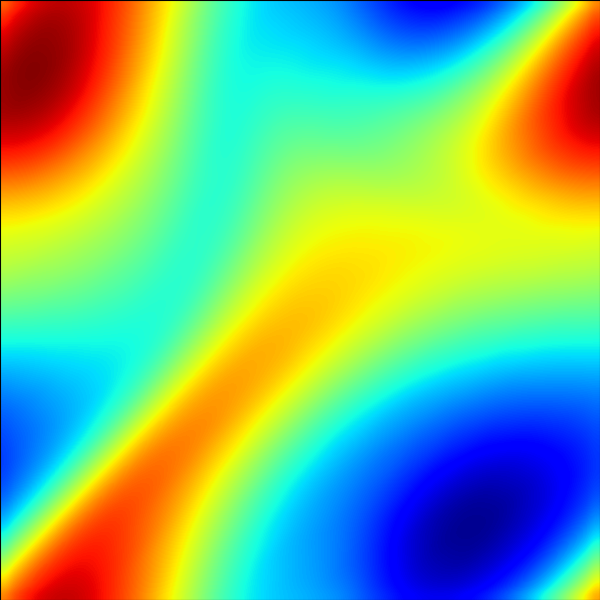}
        \includegraphics[valign=c]{Figures/ex_article-figure13.pdf}
        \includegraphics[width=0.2\textwidth, valign=c]{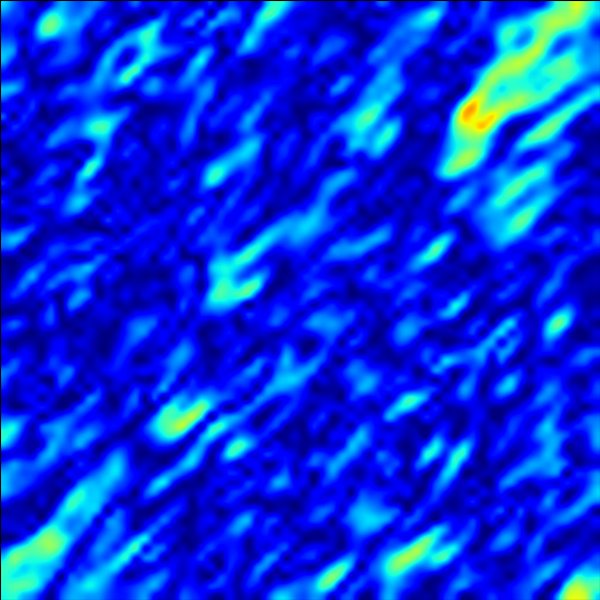}
        \includegraphics[valign=c]{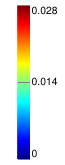}
        \vspace{0.2cm}
        \includegraphics[width=0.2\textwidth, valign=c]{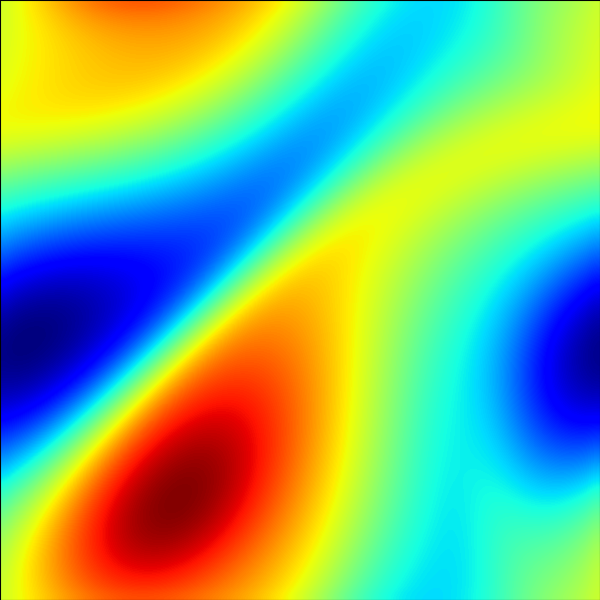}
        \includegraphics[valign=c]{Figures/ex_article-figure13.pdf}
        \includegraphics[width=0.2\textwidth, valign=c]{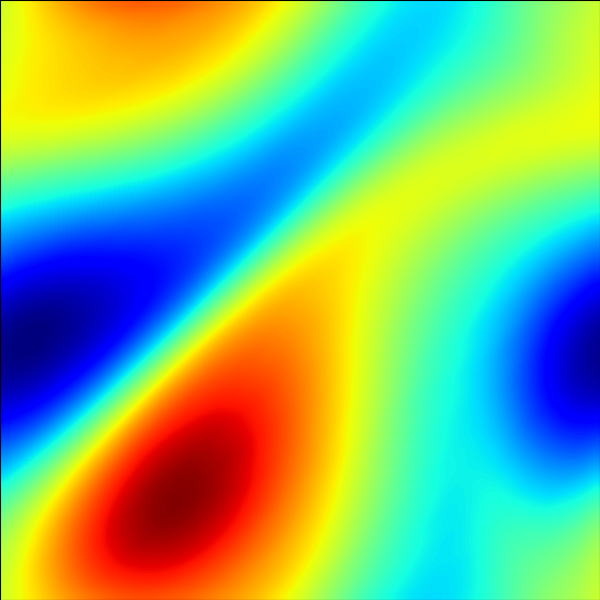}
        \includegraphics[valign=c]{Figures/ex_article-figure13.pdf}
        \includegraphics[width=0.2\textwidth, valign=c]{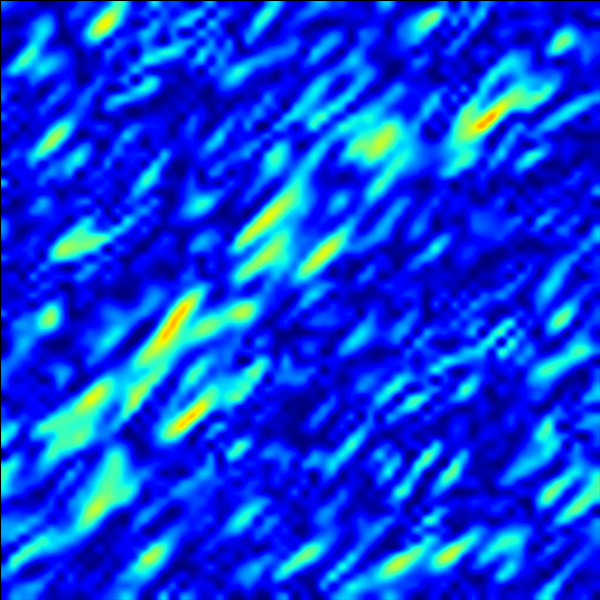}
        \includegraphics[valign=c]{Figures/ex_article-figure15.pdf}
    \caption{Two-dimensional Navier--Stokes equations. Vorticity at $t=60$ (top row) and $100$ (bottom row) predicted by the model trained with the GDSG method.}
    \label{fig:ns}
\end{figure}

\begin{figure}[thb!]
    \centering
        \includegraphics[width=0.48\linewidth, valign=t]{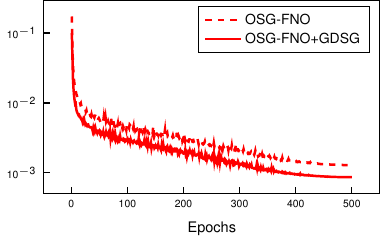}
        \includegraphics[width=0.4825\linewidth, valign=t]{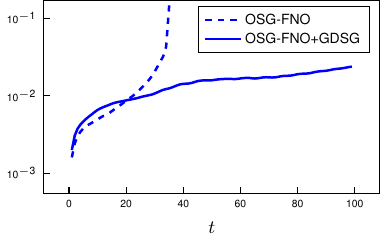}
\caption{Two-dimensional Navier--Stokes equations. Left: training loss recorded after every epoch. Right: average relative $\ell_2$ error computed on the test set with $100$ new and unseen trajectories with a length of $100$ and time lag $\Delta=1$.}
    \label{fig:ns_err}
\end{figure}

This example illustrates learning the incompressible Navier--Stokes equations \cite{li2021fourier,chen2023deep}:
\begin{equation}
	\begin{cases}
		\partial_t \omega (x,t) + {\bf v}(x,t) \cdot \nabla \omega (x,t) = \nu \Delta \omega(x,t) + f(x), \quad  & x\in (0,1)^2, t> 0,\\
		\nabla \cdot {\bf v}(x,t) = 0, & x\in (0,1)^2, t>0,\\
		\omega (x,t=0) = \omega_0(x),& x\in (0,1)^2,
	\end{cases}
	\label{eq:ns}
\end{equation}
where ${\bf v}(x,t)$ is the velocity, $\omega = \nabla \times {\bf v}$ is the vorticity, $\nu=10^{-3}$ denotes viscosity, and $f(x)=0.1(\sin (2\pi(x_1+x_2)) + \cos (2\pi(x_1+x_2)))$ represents a periodic external force. Our goal is to learn the evolution operators of $\omega$ from data with varied time lags. We use the data from  \cite{chen2023deep}, which comprises $N=100$ trajectories of solution snapshots with a length of $50$. Solutions are sampled on a $64 \times 64$ uniform grid, with time lags randomly sampled from the uniform distribution on $[0.5,1.5]$.
For neural network modeling, we construct an OSG-Net with the Fourier neural operator as the basic block, implemented as {\tt \textbf{osg\_fno}} in DUE. The two hyperparameters $\lambda$ and $Q$ are both set to $1$ for the GDSG loss function. We train the model for $500$ epochs with a batch size of $20$. Subsequently, the trained model is evaluated on $100$ new and unseen trajectories with a length of $100$ and a time step size $\Delta=1$. As shown in \cref{fig:ns}, the model trained with the GDSG method produces accurate predictions at $t=60$ and $100$. \cref{fig:ns_err} displays the training loss and testing errors. We observe that the purely data-driven model, which is trained using only the plain fitting loss without embedding the semigroup property, is unstable.


\subsection{Two-dimensional Flow Past a Circular Cylinder}
\label{sec:cylinder}

\begin{figure}[thb!]
\centering
    \includegraphics[width=1.0\linewidth]{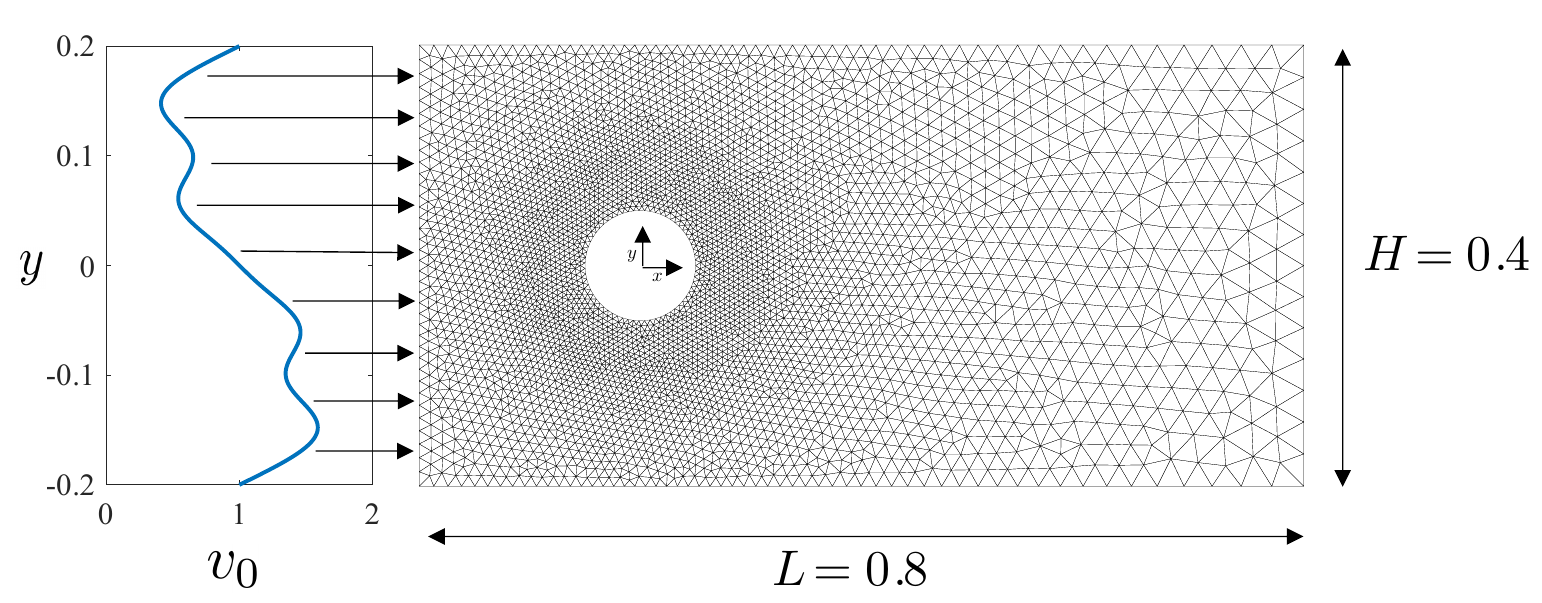}
    \caption{Two-dimensional flow past a circular cylinder at the origin in a rectangular domain. The inlet is $0.2$ units upstream of the cylinder's centroid. Domain size: $0.8$ width, $0.4$ height. Inflow has zero vertical velocity. Lateral boundaries: $\textbf{v}=(1,0)$. Outflow: zero pressure. No-slip condition on the cylinder's surface.}
    \label{fig:cylinder_setup}
\end{figure}

In this classic fluid mechanics example, we use DUE to learn the dynamics of fluid velocity $\bf v$ and pressure $p$ around a circular cylinder, generating periodic oscillations at a low Reynolds number. Synthetic data is generated by numerically solving the incompressible Navier--Stokes equations:
\begin{equation}
	\begin{cases}
		\partial_t {\bf v} + {\bf v} \cdot \nabla {\bf v} = -\frac{1}{\rho}\nabla p+\nu \Delta {\bf v}, \quad  & x\in \Omega, t> 0,\\
		\nabla \cdot {\bf v}(x,t) = 0, & x\in \Omega, t>0,
	\end{cases}
	\label{eq:ns_cylinder}
\end{equation}
with fluid density $\rho=1$ and viscosity $\nu=0.001$. The geometric configuration, boundary conditions, and computing mesh are depicted in \Cref{fig:cylinder_setup}. The horizontal velocity component at the inlet, denoted as $v_0$, is sampled from the following Fourier series with random coefficients:
\begin{equation}
	\label{eq:init_Cylinder}
	v_0(y,t)=1+0.6 \sum_{m=1}^{5} a_m\sin \left(\frac{2m\pi}{H}y\right),
\end{equation}
where $a_m\sim U[-1/m,1/m]$, and $H$ is the height of the rectangular domain.

\begin{figure}[thb!]
\centering
        \includegraphics[width=0.24\textwidth, valign=c]{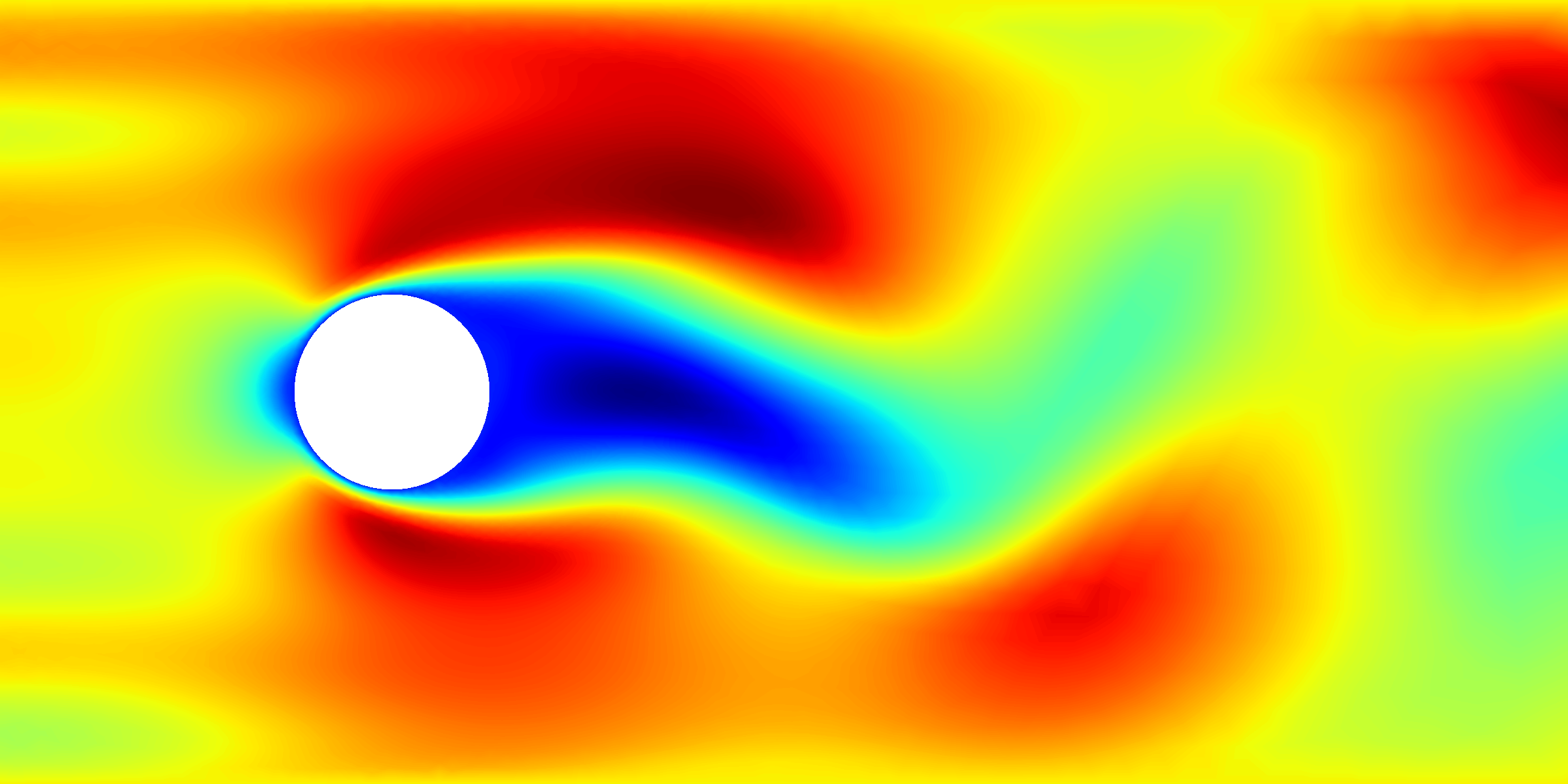}
        \includegraphics[width=0.24\textwidth, valign=c]{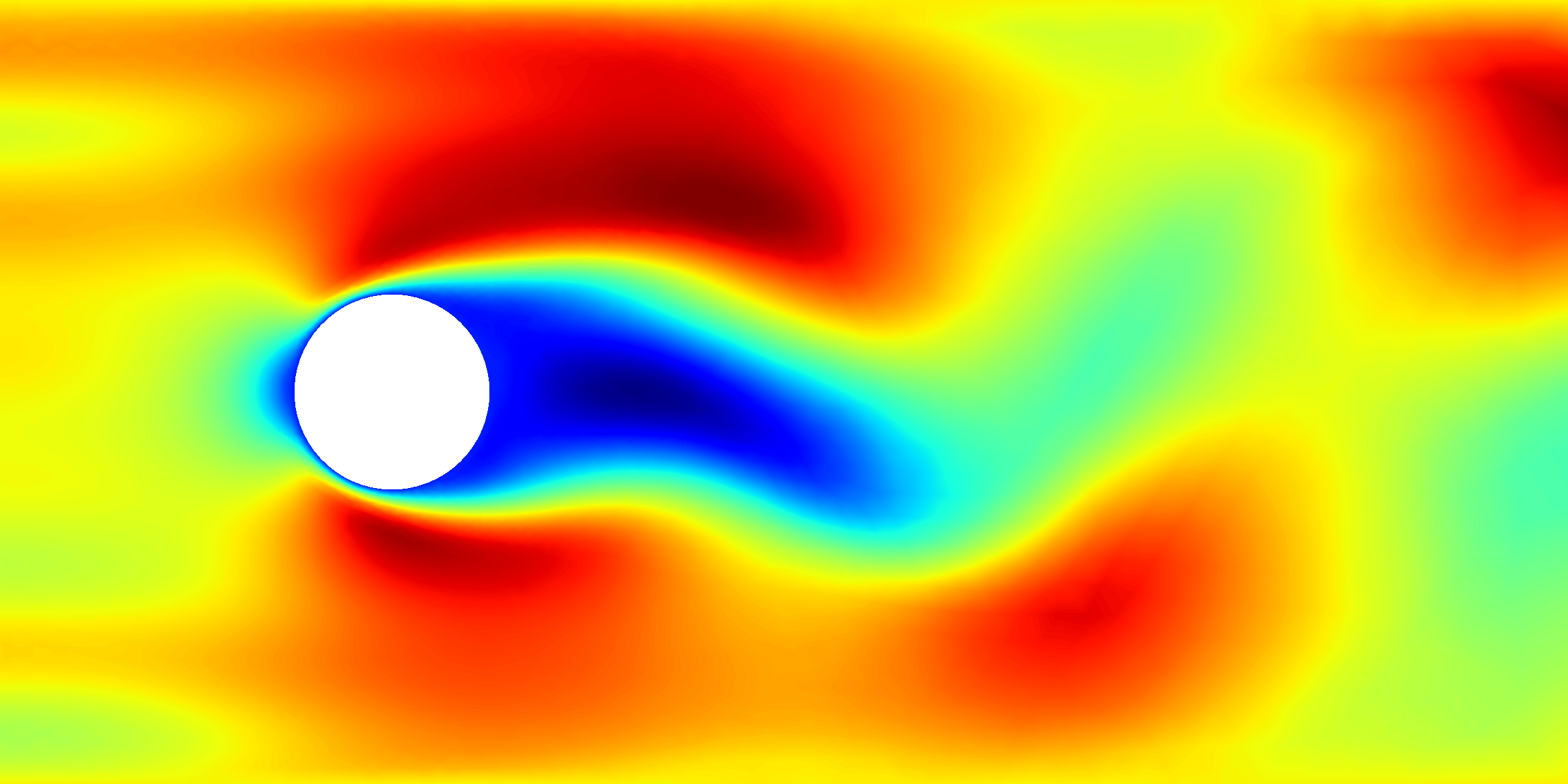}
        \includegraphics[valign=c]{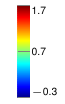}
        \includegraphics[width=0.24\textwidth, valign=c]{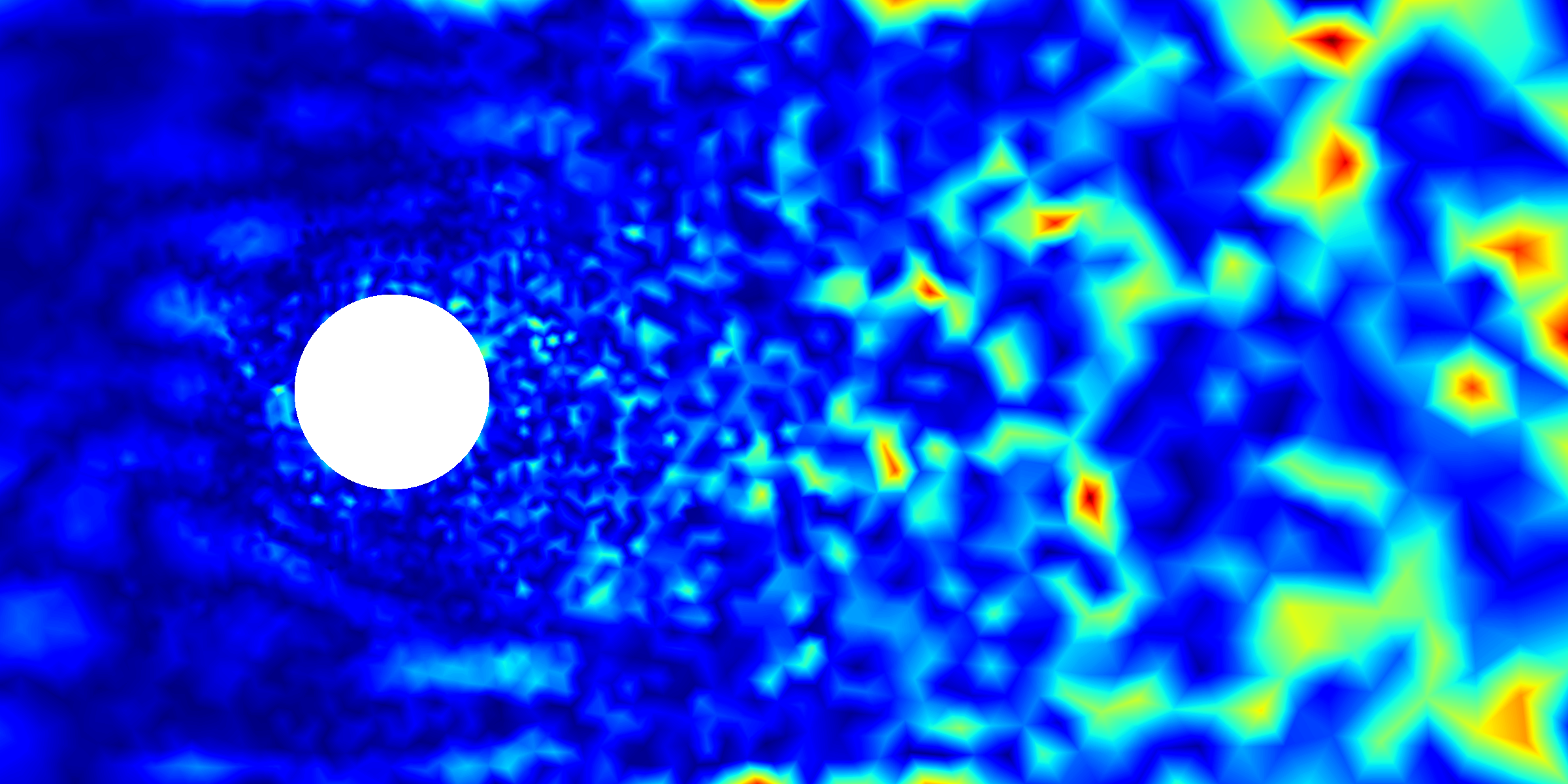}
        \includegraphics[valign=c]{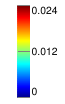}
        \includegraphics[width=0.24\textwidth, valign=c]{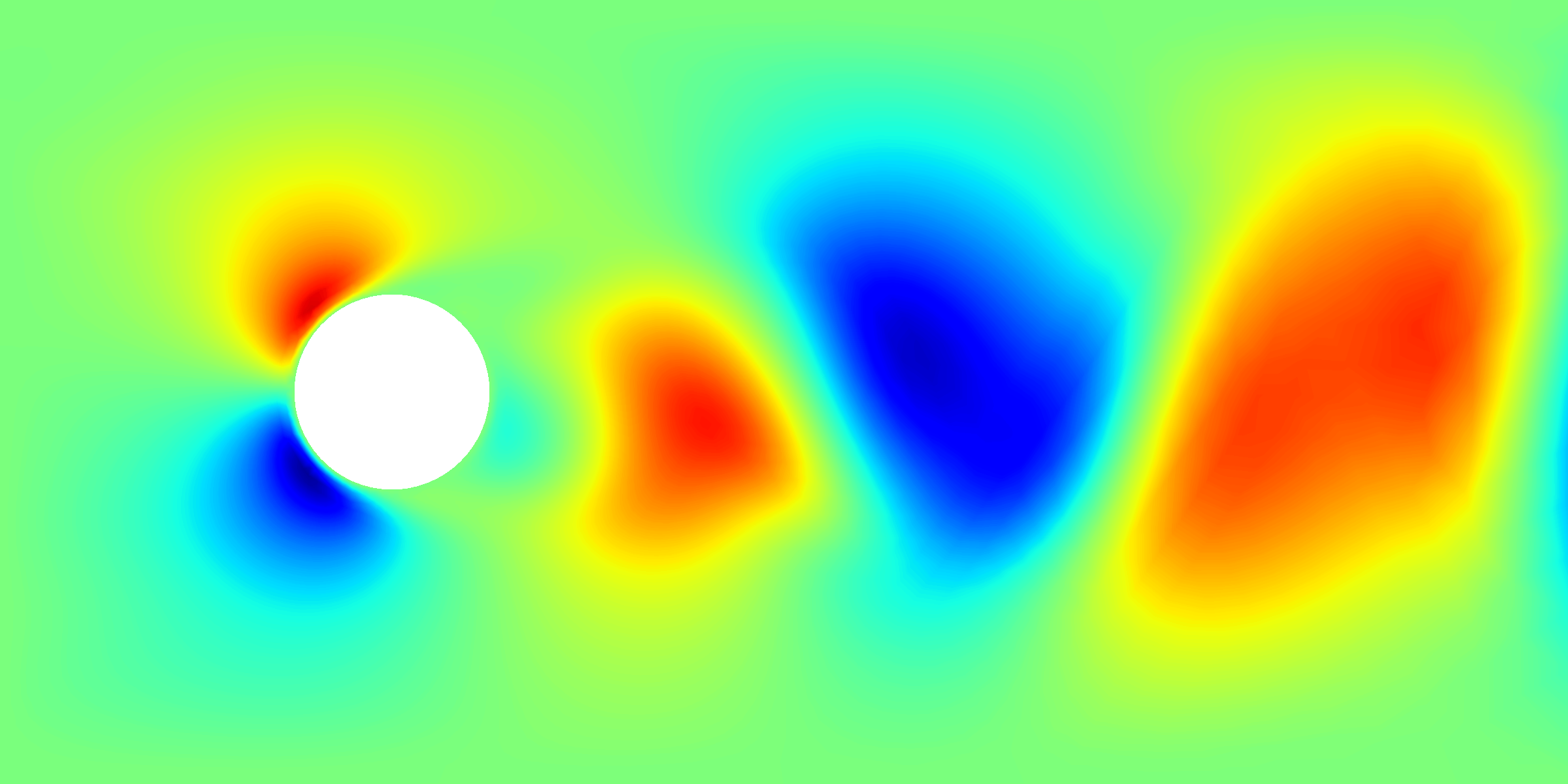}
        \includegraphics[width=0.24\textwidth, valign=c]{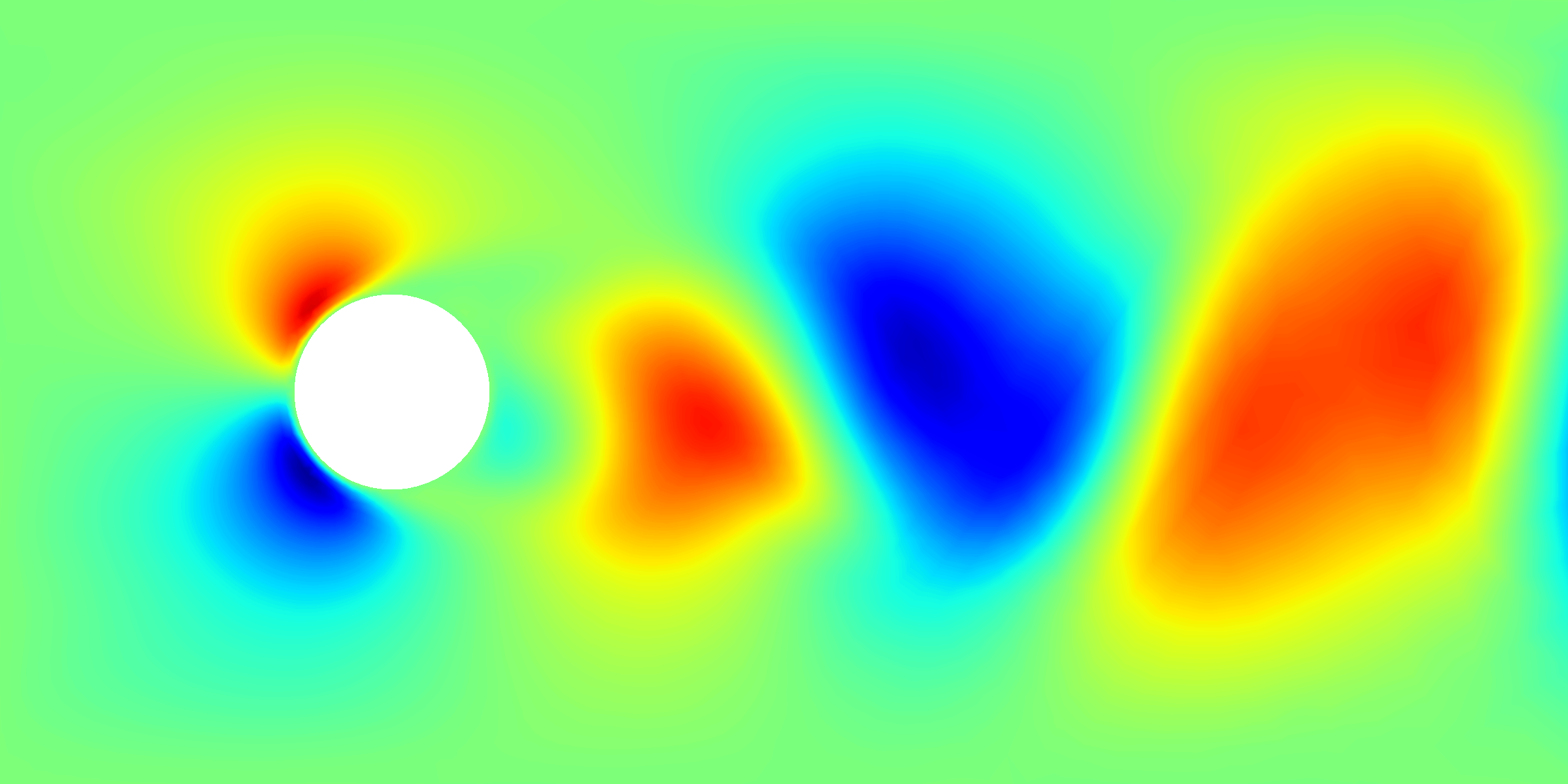}
        \includegraphics[valign=c]{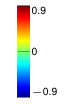}
        \includegraphics[width=0.24\textwidth, valign=c]{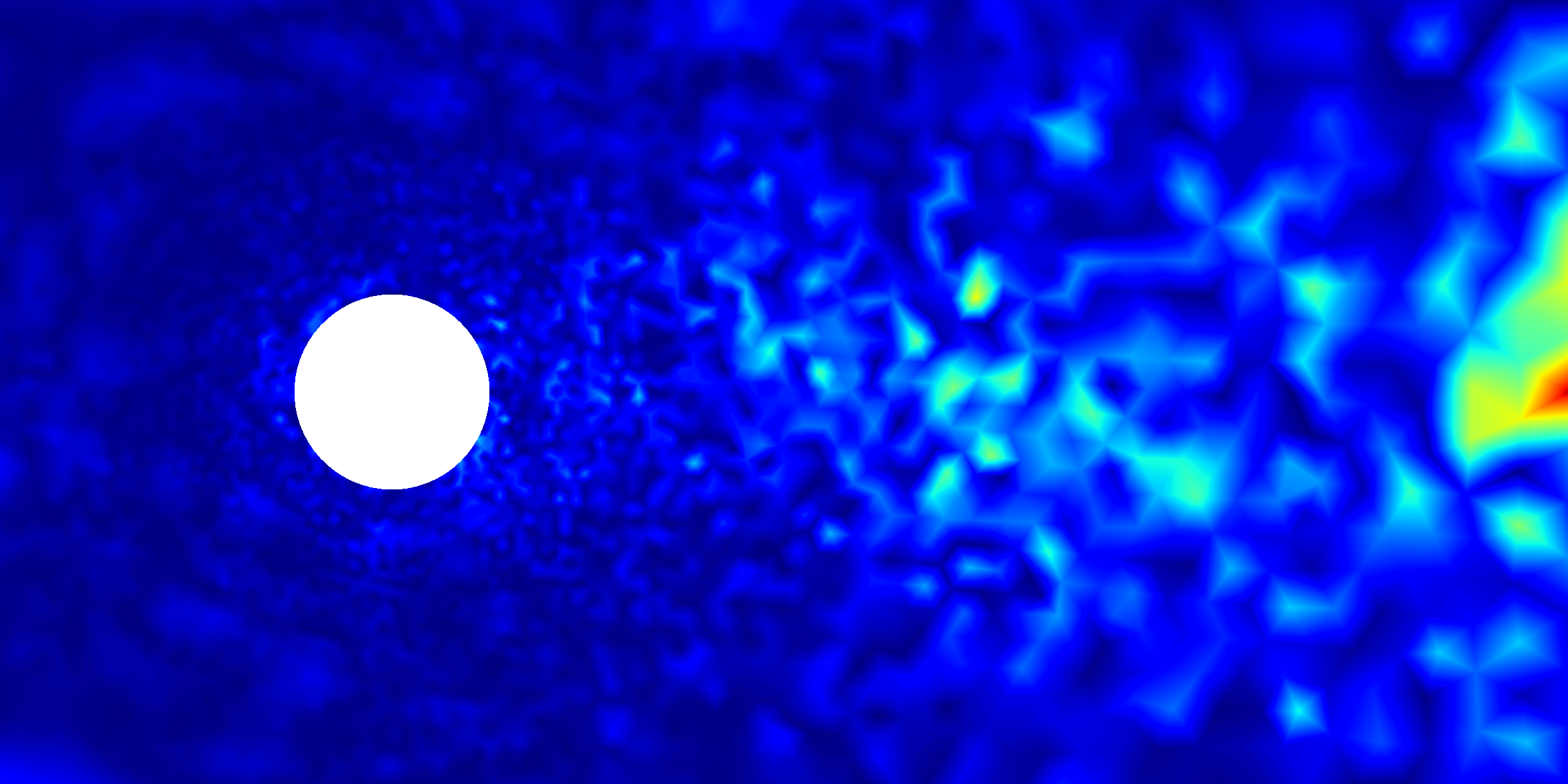}
        \includegraphics[valign=c]{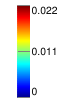}
        \vspace{0.2cm}
        \includegraphics[width=0.24\textwidth, valign=c]{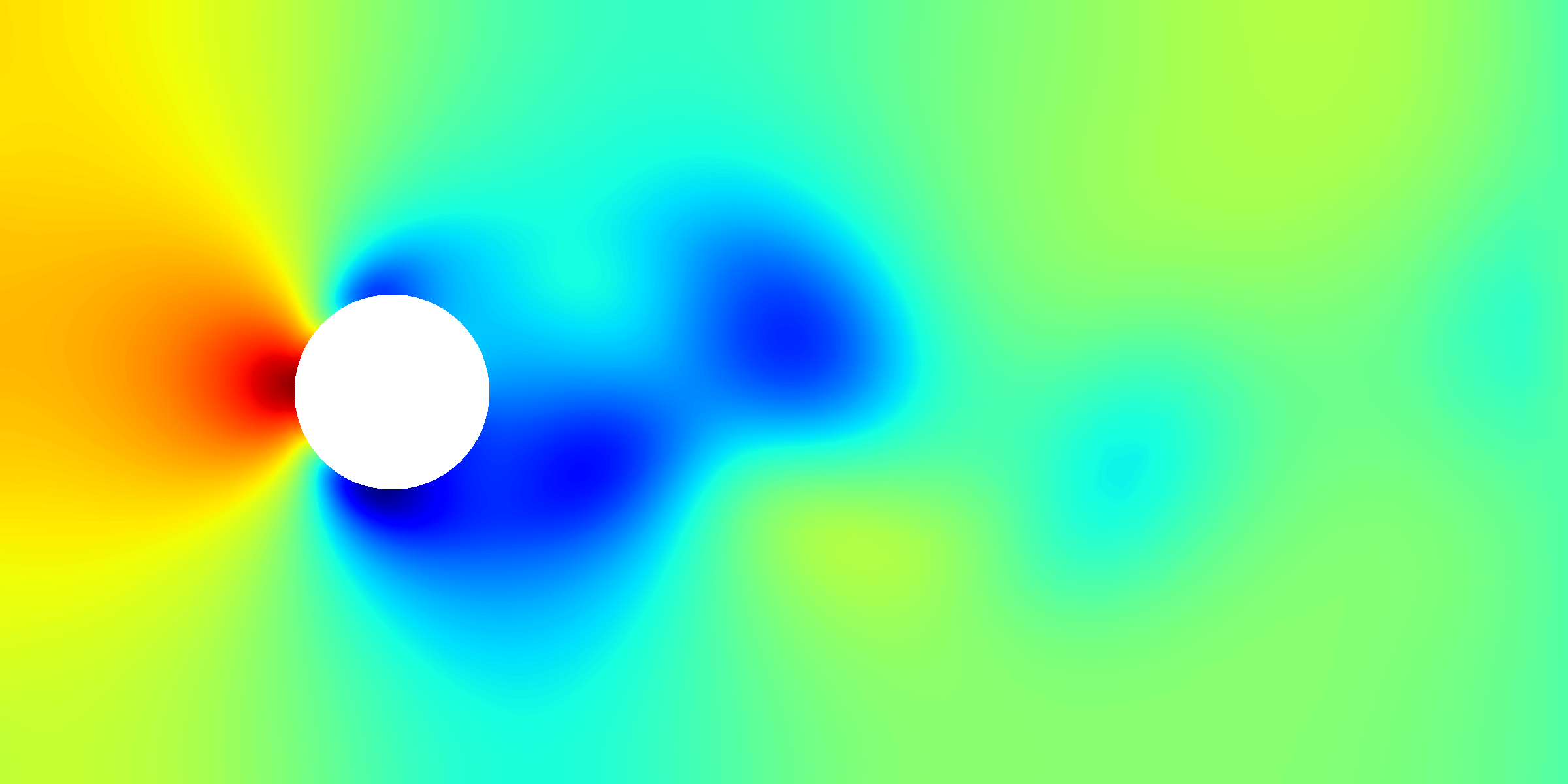}
        \includegraphics[width=0.24\textwidth, valign=c]{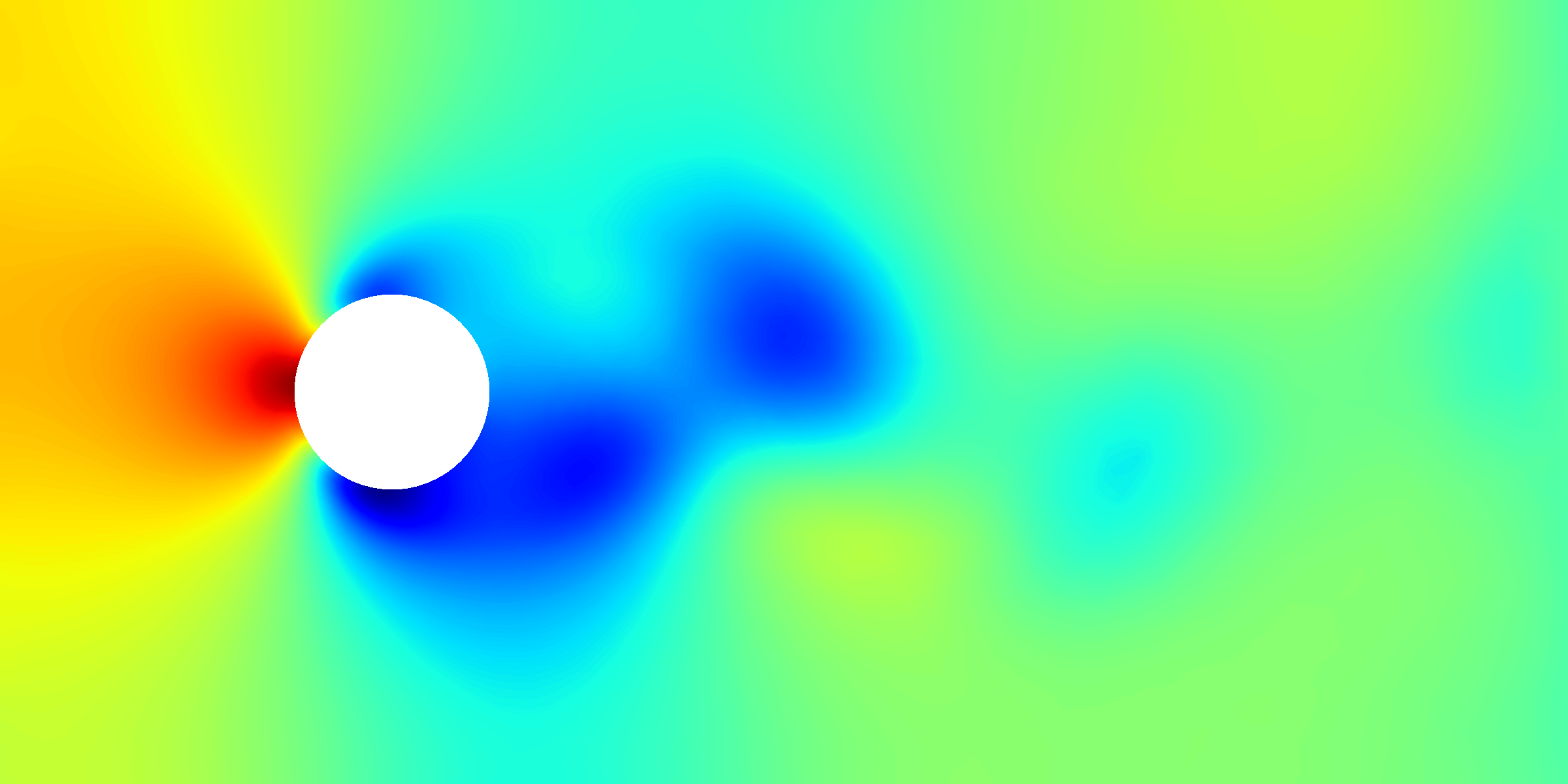}
        \includegraphics[valign=c]{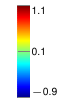}
        \includegraphics[width=0.24\textwidth, valign=c]{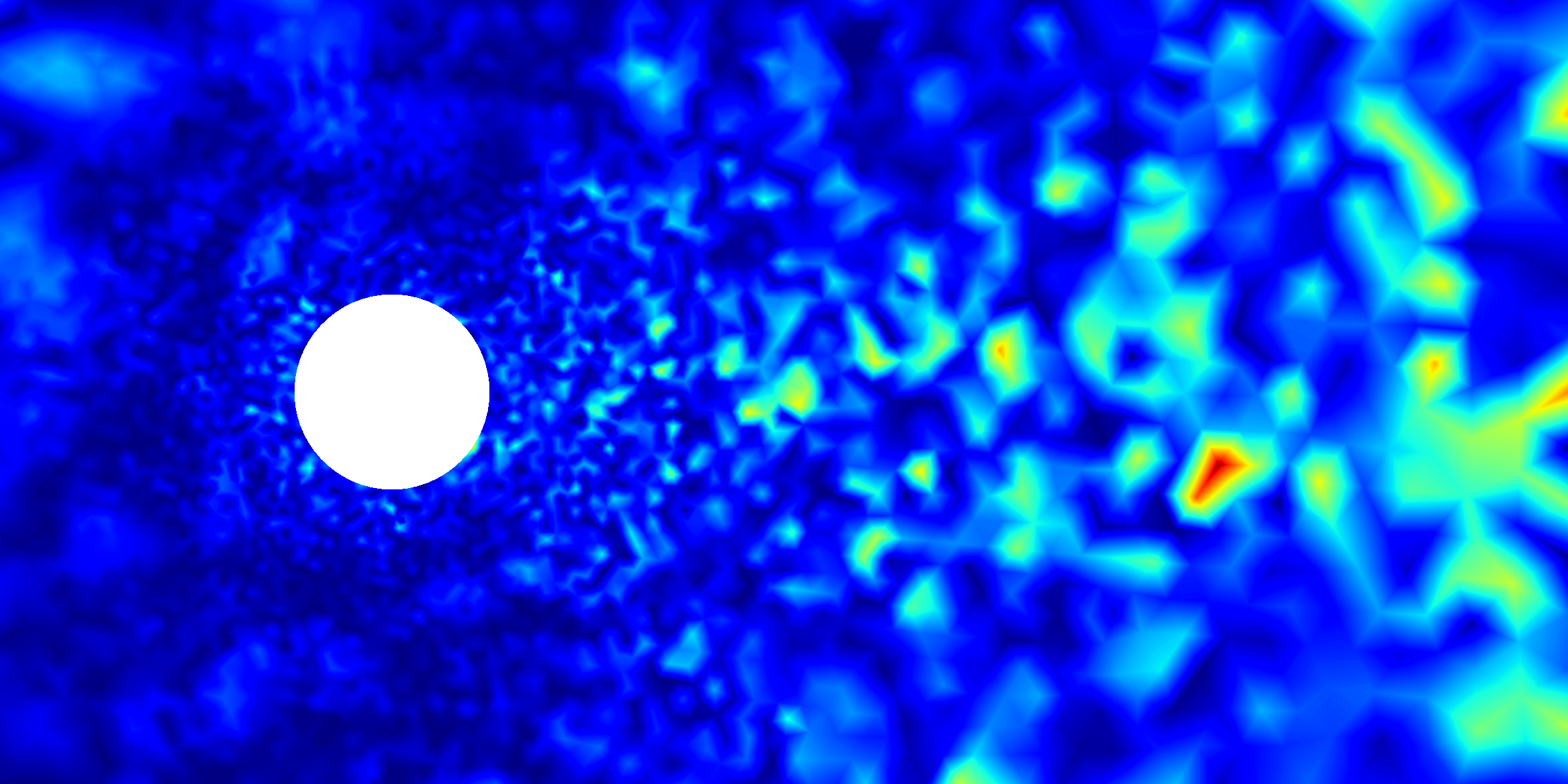}
        \includegraphics[valign=c]{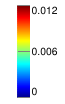}
    \caption{Two-dimensional flow past a circular cylinder. From top to bottom: horizontal velocity; vertical velocity; pressure. Left: the referential $(\textbf{v},p)$ at $t=0.5$; Middle: the predicted $(\textbf{v},p)$ given by the PiT model; Right: the absolute errors between the references and the predictions.}
    \label{fig:pred_cylinder}
\end{figure}

The dataset consists of 1,000 trajectories with 11 snapshots each, having a time lag of $0.05$. We set $K=0$ for the multi-step loss and rearrange each trajectory into 10 input-output pairs to construct the training data set. For neural network modeling with data sampled on an unstructured mesh, we employ a ResNet with the Position-induced Transformer (PiT) as the basic block, implemented as {\tt \textbf{pit}} in DUE. The model is trained for 500 epochs with a batch size of 50. Subsequently, the trained model is evaluated on 100 new and unseen trajectories with 10 forward time steps (up to $t=0.5$). As shown in \Cref{fig:pred_cylinder}, the PiT model in DUE successfully captures the dynamics of both the velocity and pressure. The relatively larger error in the downstream region is due to a sparser distribution of sampling grid points, resulting in a lower resolution of the downstream flow. Consequently, this region contributes less to the loss function, leading the model to learn less about the flow patterns there. The resolution in this region can be improved by locally increasing the number of sampling points.

%% file: 5_Conclusion.tex
\section{Conclusions and Prospects}
\label{sec:conclusion}

Artificial intelligence is revolutionizing scientific research, offering profound insights and accelerating discoveries across various fields through advanced data analysis and predictive modeling. This paper has introduced a comprehensive framework for learning unknown equations using deep learning, featuring advanced neural network architectures such as ResNet, gResNet, OSG-Net, and Transformers. This adaptable framework is capable of learning unknown ODEs, PDEs, DAEs, IDEs, and SDEs, as well as reduced or partially observed systems with missing variables. Compared to DMD, which offers faster training times and performs well on linear systems, the deep learning framework requires more computational resources for training but excels at capturing nonlinear interactions and modeling complex systems, providing greater flexibility and accuracy for tackling challenging problems.

We have presented the novel dual-OSG-Net architecture to address the challenges posed by multi-scale stiff differential equations, enabling accurate learning of dynamics across broad time scales. Additionally, we have  introduced several techniques to enhance prediction accuracy and stability, including a multi-step loss function that considers model predictions several steps ahead during training, and a semigroup-informed loss function that embeds the semigroup property into the models. These techniques could serve as examples for students and newcomers, illustrating the frontier of embedding prior knowledge into deep learning for data-driven discovery and developing structure-preserving AI for modeling unknown equations.

To support this framework, we developed Deep Unknown Equations (DUE), a user-friendly, comprehensive software tool equipped with extensive functionalities for modeling unknown equations through deep learning. DUE facilitates rapid scripting, allowing users to initiate new modeling tasks with just a few lines of code. It serves as both an educational toolbox for students and newcomers and a versatile Python library for researchers dealing with differential equations. DUE is applicable not only for learning unknown equations from data but also for surrogate modeling of known, yet complex, equations that are costly to solve using traditional numerical methods. The extensive numerical examples presented in this paper demonstrate DUE's power in modeling unknown equations, and the source codes for these examples are available in our GitHub repository, providing templates that users can easily adapt for their research.

Looking ahead, DUE is envisioned as a long-term project with ongoing maintenance and regular updates to incorporate advanced techniques. We are committed to continuously optimizing DUE’s performance and adding new functionalities as research in this field progresses. We also encourage contributions from users to expand DUE’s capabilities and broaden its applicability across a wider range of scenarios. 
One promising direction is to implement robust denoising procedures during data preprocessing, enabling DUE to achieve reliable results even with high levels of noise in the data. Additionally, reducing the amount of data required for effective deep learning performance is valuable. While the current semigroup-informed learning approach helps in this regard, incorporating additional physical constraints or leveraging prior models and knowledge could further guide the model toward accurate predictions with less data.  
Another effective strategy is active learning, which focuses on selecting the most informative data points for model training. By concentrating on critical data, active learning can enhance model performance while reducing data requirements. Lastly, transfer learning offers a powerful approach to minimize data needs further by utilizing pre-trained models on related tasks. For instance, neural operators, with their discretization-invariant properties, can be pre-trained on coarser data and adapted to finer resolutions with minimal or no retraining. Exploring additional transfer learning techniques, such as those tailored to multi-frequency time series data, is also a promising direction.

	\section*{Acknowledgment}
	The authors would like to express their sincere gratitude to the anonymous reviewers for their insightful comments and constructive suggestions, which have enhanced the quality of this paper.